\documentclass{article}
\usepackage[T1]{fontenc}    

\usepackage[numbers]{natbib}

\usepackage{adjustbox}
\usepackage{caption}
\usepackage{microtype}      
\usepackage[utf8]{inputenc} 
\usepackage{hyperref}       
\usepackage{url}            
\usepackage{booktabs}       
\usepackage{amsfonts}       
\usepackage{nicefrac}       
\usepackage{tabularx}       
\usepackage{makecell}       
\usepackage{ragged2e}       
\usepackage{paralist}       
\usepackage{arydshln}       

\usepackage{import}         
\usepackage{standalone}     
\usepackage{titlesec}       
\usepackage{multirow}
\usepackage{enumitem}
  \newlist{inlinelist}{enumerate*}{1}
  \setlist*[inlinelist,1]{%
          label=(\roman*),
      }
\usepackage{amsmath}        
\usepackage{cases}          
\usepackage{amsthm}         
\usepackage{thmtools}       
\usepackage{mathtools}      
\usepackage{subcaption}     
\usepackage{floatrow}       
\usepackage{pgf}
\usepackage{tikz}
\usepackage{pgfplots}
	\usepgfplotslibrary{groupplots}
  \usetikzlibrary{calc}
  \usetikzlibrary{positioning}
  \usetikzlibrary{angles,quotes}
  \usetikzlibrary{backgrounds}
  \usetikzlibrary{external}
  \usetikzlibrary{fit}
  \usetikzlibrary{bayesnet}
  \usetikzlibrary{calc}
  \usetikzlibrary{fit}
  \usetikzlibrary{spy, shadows}
  \usetikzlibrary{arrows.meta}
  \usetikzlibrary{shadings}
  \usetikzlibrary{fadings}
\usetikzlibrary{matrix}     
\usepackage[
    noabbrev,
    capitalize,
    nameinlink]{cleveref}   

\begingroup
    \makeatletter
    \@for\theoremstyle:=theorem,corollary,lemma,model,definition,remark,plain\do{%
        \expandafter\g@addto@macro\csname th@\theoremstyle\endcsname{%
            \setlength\thm@preskip\parskip
            \setlength\thm@postskip\parskip
            \addtolength\thm@preskip\parskip
            }%
        }
    \makeatother
\endgroup

\theoremstyle{definition}

\crefname{theorem}{Theorem}{Theorems}
\Crefname{theorem}{Theorem}{Theorems}
\crefname{lemma}{Lemma}{Lemmas}
\Crefname{lemma}{Lemma}{Lemmas}
\crefname{proposition}{Proposition}{Propositions}
\Crefname{proposition}{Proposition}{Propositions}
\crefname{definition}{Definition}{Definitions}
\Crefname{definition}{Definition}{Definitions}

\newcolumntype{L}{>{\RaggedRight\arraybackslash}X}

\newcommand{\cnaps}{\textsc{CNAPs}}

\definecolor{mydarkblue}{rgb}{0,0.08,0.45}
\hypersetup{
    colorlinks=true,
    linkcolor=mydarkblue,
    citecolor=mydarkblue,
    filecolor=mydarkblue,
    urlcolor=mydarkblue}

\definecolor{tabgreen}{HTML}{59a14f}

\titlespacing{\section}{0pt}{.5\parskip}{.25\parskip}

\newcommand{\jradd}[1]{}
\newcommand{\jbadd}[1]{}
\newcommand{\jgadd}[1]{}
\newcommand{\retadd}[1]{}
\newcommand{\snadd}[1]{}

\usepackage{amsmath,amsfonts,bm}

\def\eqref#1{equation~\ref{#1}}

\def\1{\bm{1}}

\def\vtheta{{\bm{\theta}}}
\def\vgamma{{\bm{\gamma}}}
\def\vbeta{{\bm{\beta}}}
\def\vphi{{\bm{\phi}}}
\def\vpi{{\bm{\pi}}}
\def\vpsi{{\bm{\psi}}}

\def\vf{{\bm{f}}}

\def\vx{{\bm{x}}}
\def\vy{{\bm{y}}}
\def\vz{{\bm{z}}}

\DeclareMathAlphabet{\mathsfit}{\encodingdefault}{\sfdefault}{m}{sl}
\SetMathAlphabet{\mathsfit}{bold}{\encodingdefault}{\sfdefault}{bx}{n}

\newcommand{\R}{\mathbb{R}}

\usepackage[final]{neurips_2019}

\usepackage[utf8]{inputenc} 
\usepackage[T1]{fontenc}    
\usepackage{url}            
\usepackage{booktabs}       
\usepackage{amsfonts}       
\usepackage{nicefrac}       
\usepackage{microtype}      
\usepackage{pdflscape}
\usepackage{algorithm}      
\usepackage{algpseudocode}  

\title{Fast and Flexible Multi-Task Classification Using Conditional Neural Adaptive Processes}

\author{
 James Requeima\thanks{Authors contributed equally} \\
 University of Cambridge \\
 Invenia Labs \\
 \texttt{jrr41@cam.ac.uk} \\
 \And
 Jonathan Gordon\footnotemark[1] \\
 University of Cambridge \\
 \texttt{jg801@cam.ac.uk} \\
 \And
 John Bronskill\footnotemark[1] \\
 University of Cambridge \\
 \texttt{jfb54@cam.ac.uk} \\
 \And
 Sebastian Nowozin \\
 Google Research Berlin \\
 \texttt{nowozin@google.com}\\
 \And
 Richard E.~Turner \\
 University of Cambridge \\
 Microsoft Research \\
 \texttt{ret26@cam.ac.uk}
}

\begin{document}

\maketitle

\begin{abstract}

The goal of this paper is to design image classification systems that, after an initial multi-task training phase, can automatically adapt to new tasks encountered at test time. We introduce a conditional neural process based approach to the multi-task classification setting for this purpose, and establish connections to the meta-learning and few-shot learning literature. The resulting approach, called \cnaps{}, comprises a classifier whose parameters are modulated by an adaptation network that takes the current task's dataset as input. We demonstrate that \cnaps{} achieves state-of-the-art results on the challenging \textsc{Meta-Dataset} benchmark indicating high-quality transfer-learning. We show that the approach is robust, avoiding both over-fitting in low-shot regimes and under-fitting in high-shot regimes. Timing experiments reveal that \cnaps{} is computationally efficient at test-time as it does not involve gradient based adaptation. Finally, we show that trained models are immediately deployable to continual learning and active learning where they can outperform existing approaches that do not leverage transfer learning.

\end{abstract}

\section{Introduction}
\label{sec:intro}

We consider the development of general purpose image classification systems that can handle tasks from a broad range of data distributions, in both the low and high data regimes, without the need for costly retraining when new tasks are encountered. We argue that such systems require mechanisms that adapt to each task, and that these mechanisms should themselves be learned from a diversity of datasets and tasks at training time. This general approach relates to methods for meta-learning \citep{schmidhuber1987evolutionary,thrun2012learning} and few-shot learning \citep{lake2015human}. However, existing work in this area typically considers homogeneous task distributions at train and test-time that therefore require only minimal adaptation. To handle the more challenging case of different task distributions we design a fully adaptive system, requiring specific design choices in the model and training procedure.

Current approaches to meta-learning and few-shot learning for classification are characterized by two fundamental trade-offs.
\begin{inlinelist}
\item The number of parameters that are adapted to each task. One approach adapts only the top, or head, of the classifier leaving the feature extractor fixed \citep{snell2017prototypical,gordon2018meta}. While useful in simple settings, this approach is prone to under-fitting when the task distribution is heterogeneous \citep{triantafillou2019meta}. Alternatively, we can adapt all parameters in the feature extractor \citep{finn2017model,nichol2018reptile} thereby increasing fitting capacity, but incurring a computation cost and opening the door to over-fitting in the low-shot regime. What is needed is a middle ground which strikes a balance between model capacity and reliability of the adaptation. 
\item The adaptation mechanism. Many approaches use gradient-based adaptation \citep{finn2017model,yosinski2014transferable}. While this approach can incorporate training data in a very flexible way, it is computationally inefficient at test-time, may require expertise to tune the optimization procedure, and is again prone to over-fitting. Conversely, function approximators can be used to directly map training data to the desired parameters (we refer to this as \textit{amortization}) \citep{gordon2018meta,qiao2017few}. This yields fixed-cost adaptation mechanisms, and enables greater sharing across training tasks. However, it may under-fit if the function approximation is not sufficiently flexible. On the other hand, high-capacity function approximators require a large number of training tasks to be learned. 
\end{inlinelist}

We introduce a modelling class that is well-positioned with respect to these two trade-offs for the multi-task classification setting called Conditional Neural Adaptive Processes (\cnaps{}).\footnote{Source code available at \url{https://github.com/cambridge-mlg/cnaps}.} \cnaps{} directly model the desired predictive distribution \citep{geisser1983prediction,geisser2017predictive}, thereby introducing a \textit{conditional neural processes} (CNPs) \citep{garnelo2018conditional} approach to the multi-task classification setting. 
\cnaps{} handles varying way classification tasks and introduces a parametrization and training procedure enabling the model to \textit{learn to adapt} the feature representation for classification of diverse tasks at test time.
\cnaps{} utilize i) a classification model with shared global parameters and a small number of task-specific parameters. We demonstrate that by identifying a small set of key parameters, the model can balance the trade-off between flexibility and robustness. ii) A rich adaptation neural network with a novel auto-regressive parameterization that avoids under-fitting while proving easy to train in practice with existing datasets \citep{triantafillou2019meta}. 
In \cref{sec:experiments} we evaluate \cnaps{}. Recently, \citet{triantafillou2019meta} proposed \textsc{Meta-Dataset}, a few-shot classification benchmark that addresses the issue of homogeneous train and test-time tasks and more closely resembles real-world few-shot multi-task learning. Many of the approaches that achieved excellent performance on simple benchmarks struggle with this collection of diverse tasks. In contrast, we show that \cnaps{} achieve state-of-the-art performance on the \textsc{Meta-Dataset} benchmark, often by comfortable margins and at a fraction of the time required by competing methods. Finally, we showcase the versatility of the model class by demonstrating that \cnaps{} can be applied ``out of the box'' to continual learning and active learning.




\ifstandalone
\newpage
\bibliography{bibliography}
\bibliographystyle{abbrvnat}
\fi

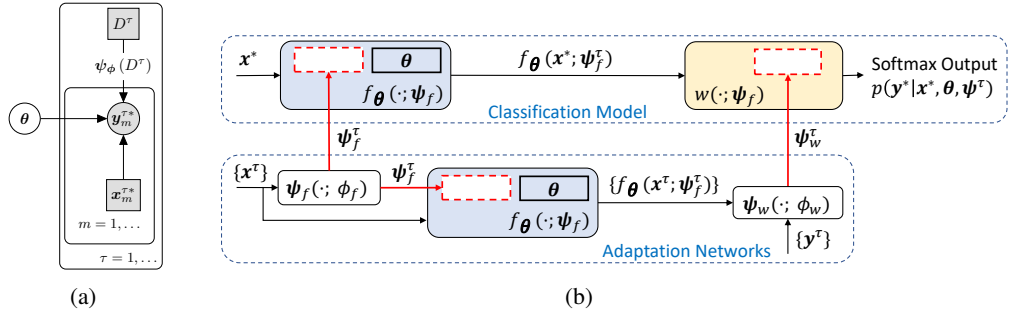
\begin{figure}[t]
	\centering
    \hspace{-20mm}
 	\begin{subfigure}[b]{.3\textwidth}
 		\centering
        \newcommand{\cons}{%
\begin{tikzpicture}[cell/.style={draw,shape=rectangle,minimum size=0.4cm}]
\begin{scope}
\node[cell](a)[anchor=west]{};
\node[cell](b)[right=0mm of a.east,anchor=west]{};
\end{scope}
\end{tikzpicture}
}

\begin{tikzpicture}
\node[obs, rectangle, inner sep=.3em,draw] (x) {$\vx^{\tau\ast}_m$};%
\node[obs, above=of x] (y) {$\vy^{\tau\ast}_m$};%
\node[above=.5 of y] (psi) {$\vpsi_{\vphi} \left( D^\tau \right)$}; %
\node[obs, rectangle, above=.3 of psi] (D) {$D^\tau$};%
\node[latent, left=1.5 of y, inner sep=.3em] (theta) {$\vtheta$}; %

\plate [inner sep=.3cm, yshift=.02cm] {plate1} {(x) (y)} {$m=1,\hdots$}; %
\plate [inner sep=.1cm] {plate3} {(plate1) (psi) (D)} {$\tau=1,\hdots$}; %

\edge {psi, x, theta} {y}; %
\draw (D) edge (psi); %
\end{tikzpicture} \\
        \subcaption{}
 		\label{fig:graphical_model}
 	\end{subfigure} %
    \hspace{-5mm}
 	\begin{subfigure}[b]{.69\textwidth}
 		\centering
        \includegraphics[width=1.1\textwidth]{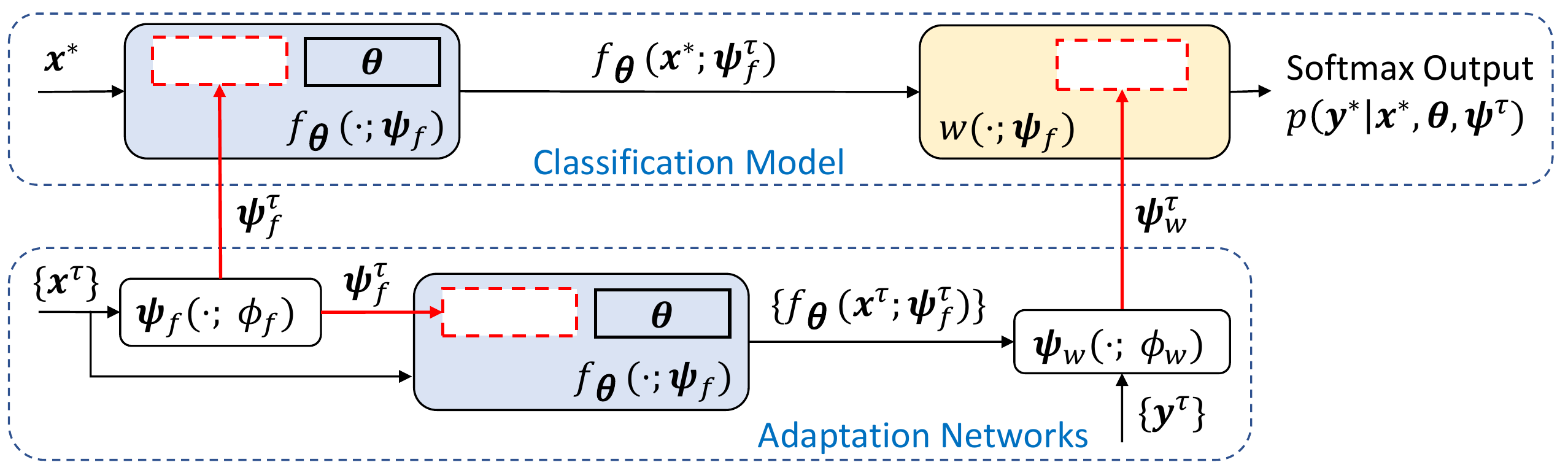} \\
        \subcaption{}
 		\label{fig:architexture}
 	\end{subfigure}
 	\caption{(a) Probabilistic graphical model detailing the CNP \citep{garnelo2018conditional} framework. (b) Computational diagram depicting the \cnaps{} model class. Red boxes imply parameters in the model architecture supplied by adaptation networks. Blue shaded boxes depict the feature extractor and the gold box depicts the linear classifier.}
 	\label{fig:graph_and_architecture}
\end{figure}

\section{Model Design}
\label{sec:model}
We consider a setup where a large number of training tasks are available, each composed of a set of inputs $\vx$ and labels $\vy$. The data for task $\tau$ includes a \textit{context set} $D^\tau=\{(\vx^\tau_n, \vy^\tau_n)\}_{n=1}^{N_\tau}$, with inputs and outputs observed, and a \textit{target set} $\{(\vx^{\tau\ast}_m, \vy^{\tau\ast}_m)\}_{m=1}^{M_\tau}$ for which we wish to make predictions ($\vy^{\tau\ast}$ are only observed during training). CNPs \citep{garnelo2018conditional} construct predictive distributions given $\vx^\ast$ as:
\begin{equation}
    p \left(\vy^\ast | \vx^\ast, \vtheta, D^\tau \right) = p \left(\vy^\ast | \vx^\ast, \vtheta, \vpsi^\tau = \vpsi_{\vphi} \left( D^\tau \right) \right).
\end{equation}
Here $\vtheta$ are global classifier parameters shared across tasks. $\vpsi^\tau$ are local task-specific parameters, produced by a function $\vpsi_{\vphi}(\cdot)$ that acts on $D^\tau$. $\vpsi_{\vphi}(\cdot)$ has another set of global parameters $\vphi$ called \textit{adaptation network parameters}. $\vtheta$ and $\vphi$ are the learnable parameters in the model (see \cref{fig:graphical_model}).

\cnaps{} is a model class that specializes the CNP framework for the multi-task classification setting. The model-class is characterized by a number of design choices, made specifically for the multi-task image classification setting. \cnaps{} employ global parameters $\vtheta$ that are trained offline to capture high-level features, facilitating transfer and multi-task learning. Whereas CNPs define $\vpsi^\tau$ to be a fixed dimensional vector used as an input to the model, \cnaps{} instead let $\vpsi^\tau$ be specific parameters of the model itself. This increases the flexibility of the classifier, enabling it to model a broader range of input / output distributions. We discuss our choices (and associated trade-offs) for these parameters below. Finally, \cnaps{} employ a novel auto-regressive parameterization of $\vpsi_{\vphi}(\cdot)$ that significantly improves performance. An overview of \cnaps{} and its key components is illustrated in \cref{fig:architexture}.

\subsection{Specification of the classifier: global \texorpdfstring{$\vtheta$}{TEXT} and task-specific parameters \texorpdfstring{$\vpsi^\tau$}{TEXT}}

\label{sec:theta_and_psi}

We begin by specifying the classifier's global parameters $\vtheta$ followed by how these are adapted by the local parameters $\vpsi^\tau$.

\label{sec:global_parameters}
\textbf{Global Classifier Parameters}. The global classifier parameters will parameterize a feature extractor $f_{\vtheta}(\vx)$ whose output is fed into a linear classifier, described below. A natural choice for $f_\vtheta(\cdot)$ in the image setting is a  convolutional neural network, e.g., a ResNet \citep{he2016deep}. In what follows, we assume that the global parameters $\vtheta$ are fixed and known. In \cref{sec:training} we discuss the training of $\vtheta$. 

\textbf{Task-Specific Classifier Parameters: Linear Classification Weights}. The final classification layer must be task-specific as each task involves distinguishing a potentially unique set of classes. We use a task specific affine transformation of the feature extractor output, followed by a softmax. The task-specific weights are denoted $\vpsi_w^\tau \in \R^{d_f \times C^\tau}$ (suppressing the biases to simplify notation), where $d_f$ is the dimension of the feature extractor output $f_\vtheta(\vx)$ and $C^\tau$ is the number of classes in task $\tau$.  

\textbf{Task-Specific Classifier Parameters: Feature Extractor Parameters}. A sufficiently flexible model must have capacity to adapt its feature representation $f_{\vtheta}(\cdot)$ as well as the classification layer (e.g.~compare the optimal features required for ImageNet versus Omiglot). We therefore introduce a set of local feature extractor parameters $\vpsi^\tau_f$, and denote $f_{\vtheta}(\cdot)$ the \textit{unadapted} feature extractor, and $f_{\vtheta}(\cdot ; \vpsi^\tau_f)$ the feature extractor adapted to task $\tau$.   

It is critical in few-shot multi-task learning to adapt the feature extractor in a parameter-efficient manner. Unconstrained adaptation of all the feature extractor parameters (e.g.~by fine-tuning \citep{yosinski2014transferable}) gives flexibility, but it is also slow and prone to over-fitting \citep{triantafillou2019meta}. Instead, we employ linear modulation of the convolutional feature maps as proposed by \citet{perez2018film}, which adapts the feature extractor through a relatively small number of task specific parameters. 
%
%

\begin{figure}[t]
	\centering
    \hfill
	\begin{subfigure}[b]{.2\textwidth}
		\centering
        \includegraphics[width=1.0\textwidth]{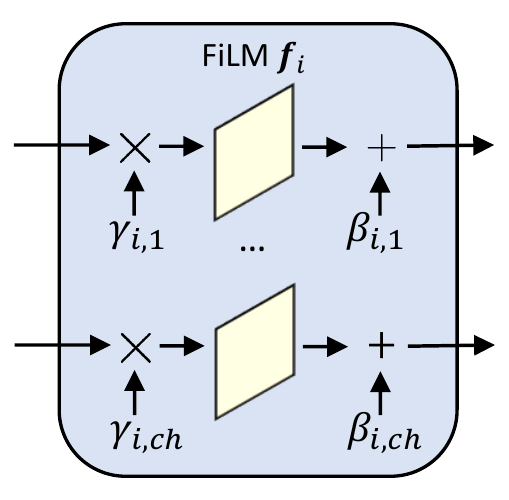}
        \subcaption{A FiLM layer.}
		\label{fig:film_layer}
	\end{subfigure} %
    \hfill
	\begin{subfigure}[b]{.78\textwidth}
		\centering
        \includegraphics[width=1.0\textwidth]{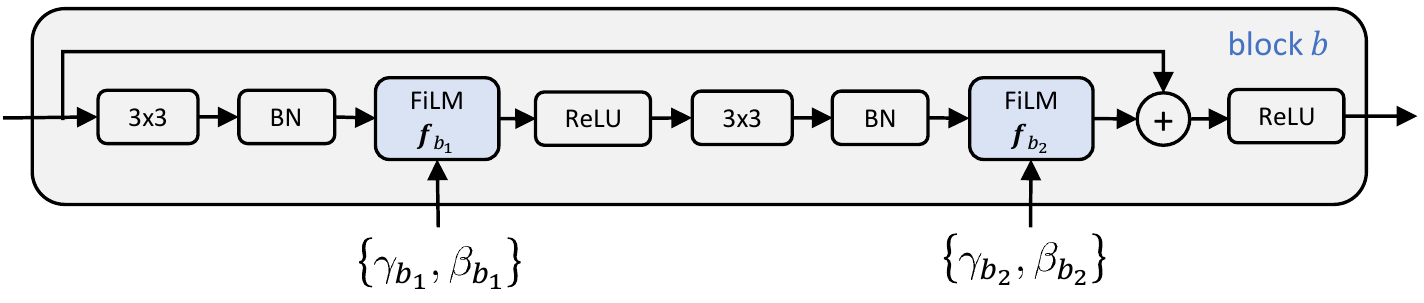}
        \subcaption{A ResNet basic block with FiLM layers.}
		\label{fig:film_res_block}
	\end{subfigure}
	\label{fig:FiLM_layer_illustrations}
	\caption{(Left) A FiLM layer operating on convolutional feature maps indexed by channel $ch$. (Right) How a FiLM layer is used within a basic Residual network block \citep{he2016deep}.}
\end{figure}

A Feature-wise Linear Modulation (FiLM) layer \citep{perez2018film} scales and shifts the $i^{th}$ unadapted feature map $\vf_{i}$ in the feature extractor $\text{FiLM}(\vf_{i}; \gamma^\tau_{i}, \beta^\tau_{i}) = \gamma^\tau_{i} \vf_{i} + \beta^\tau_{i}$ using two task specific parameters, $\gamma^\tau_{i}$ and $\beta^\tau_{i}$. \cref{fig:film_layer} illustrates a FiLM layer operating on a convolutional layer, and \cref{fig:film_res_block} illustrates how a FiLM layer can be added to a standard Residual network block \citep{he2016deep}. A key advantage of FiLM layers is that they enable expressive feature adaptation while adding only a small number of parameters \citep{perez2018film}. For example, in our implementation we use a ResNet18 with FiLM layers after every convolutional layer. The set of task specific FiLM parameters ($\vpsi^\tau_f = \{\vgamma^\tau_i, \vbeta^\tau_i\}$) constitute fewer than 0.7\% of the parameters in the model. Despite this, as we show in \cref{sec:experiments}, they allow the model to adapt to a broad class of datasets.

%


\subsection{Computing the local parameters via adaptation networks} 
\label{sec:psi_as_functions}

The previous sections have specified the form of the classifier $p\left(\vy^\ast | \vx^\ast, \vtheta, \vpsi^\tau \right)$ in terms of the global and task specific parameters,  $\vtheta$ and $\vpsi^\tau = \{ \vpsi^\tau_f, \vpsi^\tau_w \}$. The local parameters could now be learned separately for every task $\tau$ via optimization. While in practice this is feasible for small numbers of tasks (see e.g., \citep{rebuffi2017learning,rebuffi2018efficient}), this approach is computationally demanding, requires expert oversight (e.g.~for tuning early stopping), and can over-fit in the low-data regime.  


Instead, \cnaps{} uses a function, such as a neural network, that takes the context set $D^\tau$ as an input and returns the task-specific parameters, $\vpsi^\tau = \vpsi_\vphi\left( D^\tau \right)$. The adaptation network has parameters $\vphi$ that will be trained on multiple tasks to learn how to produce local parameters that result in good generalisation, a form of meta-learning. Sacrificing some of the flexibility of the optimisation approach, this method is comparatively cheap computationally (only involving a forward pass through the adaptation network), automatic (with no need for expert oversight), and employs explicit parameter sharing (via $\vphi$) across the training tasks. 
%

\begin{figure}
\centering
\includegraphics[width=1.0\textwidth]{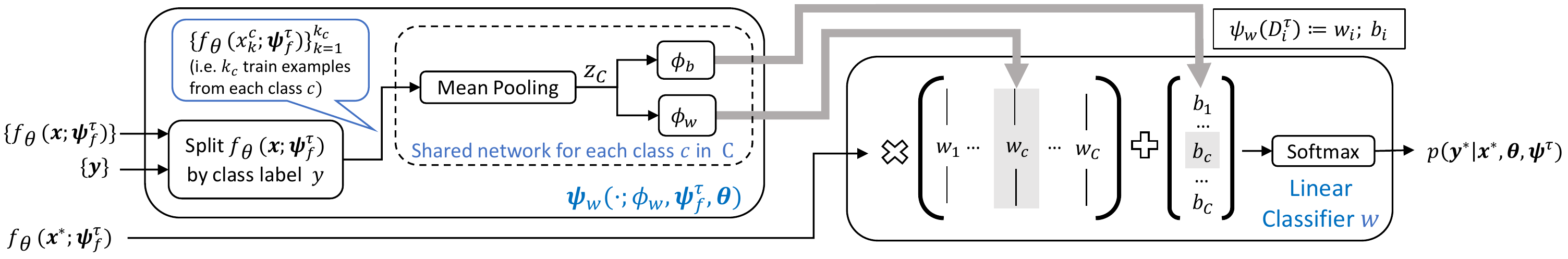}
\caption{Implementation of functional representation of the class-specific parameters $\vpsi_w$. In this parameterization, $\vpsi^c_w$  are the linear classification parameters for class $c$, and $\vphi_w$ are the learnable parameters.}
\label{fig:classifier}
\end{figure}

\label{sec:classifier_layer_amortization}
\textbf{Adaptation Network: Linear Classifier Weights}. \cnaps{} represents the linear classifier weights $\vpsi_w^\tau$ as a parameterized function of the form $\vpsi_w^\tau = \vpsi_w(D^\tau; \vphi_w, \vpsi_f, \vtheta)$, denoted $\vpsi_w(D^\tau)$ for brevity. There are three challenges with this approach: first, the dimensionality of the weights depends on the task ($\vpsi_w^\tau$ is a matrix with a column for each class, see \cref{fig:classifier}) and thus the network must output parameters of different dimensionalities; second, the number of datapoints in $D^\tau$ will also depend on the task and so the network must be able to take inputs of variable cardinality; third, we would like the model to support continual learning. To handle the first two challenges we follow \citet{gordon2018meta}. First, each column of the weight matrix is generated independently from the context points from that class $\vpsi^\tau_w = 
\begin{bmatrix}
    \vpsi_w \left( D_1^\tau \right ), & \hdots,  & \vpsi_w \left( D_C^\tau \right ) 
  \end{bmatrix}$, an approach which scales to arbitrary numbers of classes.
Second, we employ a permutation invariant architecture \citep{zaheer2017deep,qi2017pointnet} for $\vpsi_w(\cdot)$ to handle the variable input cardinality  (see \cref{app:network_architecture_details} for details). Third, as permutation invariant architectures can be incrementally updated \citep{vartak+al:2017}, continual learning is supported (as discussed in \cref{sec:experiments}).

Intuitively, the classifier weights should be determined by the representation of the  data points emerging from the adapted feature extractor. We therefore input the adapted feature representation of the data points into the network, rather than the raw data points (hence the dependency of $\vpsi_w$ on $\vpsi_f$ and $\vtheta$). To summarize, $\vpsi_w(\cdot)$ is a function \textit{on sets} that accepts as input a set of \textit{adapted} feature representations from $D_c^\tau$, and outputs the $c^{\text{th}}$ column of the linear classification matrix, i.e.,
\begin{equation}
    \vpsi_w \left(D_c^\tau; \vphi_w, \vpsi_f, \vtheta \right ) = \vpsi_w \left( \{ f_\vtheta \left( \vx_m; \vpsi_f \right) | \vx_m \in D^\tau, \vy_m=c\}; \vphi_w \right).
\end{equation}
Here $\vphi_w$ are learnable parameters of $\vpsi_w(\cdot)$. See \cref{fig:classifier} for an illustration. 

\label{sec:feature_extractor_amortization}
\textbf{Adaptation Network: Feature Extractor Parameters}.
\cnaps{} represents the task-specific feature extractor parameters $\vpsi^\tau_f$, comprising the parameters of the FiLM layers $\vgamma^\tau$ and $\vbeta^\tau$ in our implementation, as a parameterized function of the context-set $D^\tau$. Thus, $\vpsi_f(\cdot; \vphi_f, \vtheta)$ is a collection of functions (one for each FiLM layer) with parameters  $\vphi_f$, many of which are shared across functions. We denote the function generating the parameters for the $i^{\text{th}}$ FiLM layer $\vpsi^i_f(\cdot)$ for brevity.

Our experiments (\cref{sec:experiments}) show that this mapping requires careful parameterization. 
We propose a novel parameterization that improves performance in complex settings with diverse datasets. Our implementation contains two components: a task-specific representation that provides context about the task to all layers of the feature extractor (denoted $\vz_\text{G}^\tau$), and an auto-regressive component that provides information to deeper layers in the feature extractor concerning how shallower layers have adapted to the task (denoted $\vz_\text{AR}^i$). The input to the $\vpsi^i_f(\cdot)$ network is $\vz_i = (\vz_\text{G}^\tau, \vz_\text{AR}^i)$. $\vz_\text{G}^\tau$ is computed for every task $\tau$ by passing the inputs $\vx_n^\tau$ through a global set encoder $g$ with parameters in $\vphi_f$.

\begin{figure}
    \centering
    \includegraphics[width=1.0\textwidth]{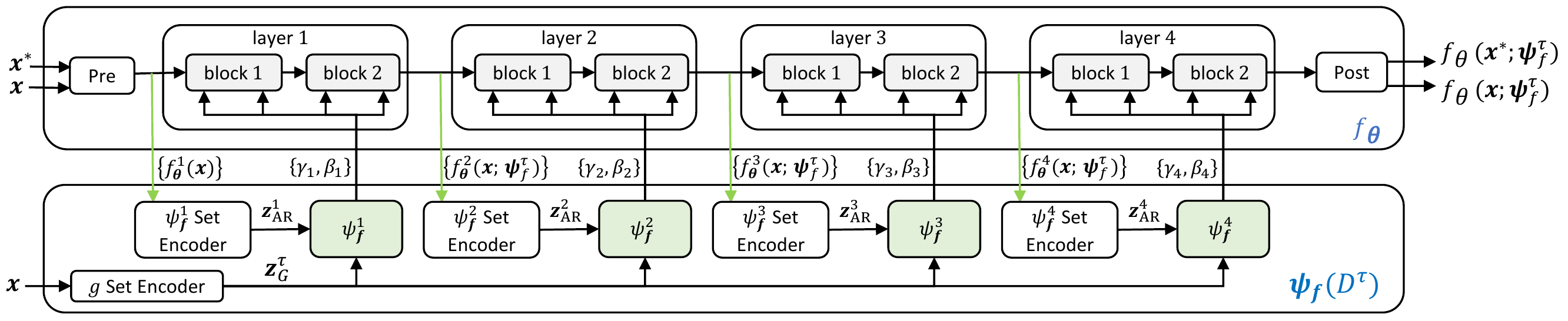}
    \caption{Implementation of the feature-extractor: an independently learned set encoder $g$ provides a fixed context that is concatenated to the (processed) activations of $\vx$ from the previous ResNet block. The inputs $\vz_i = (\vz_\text{G}^\tau, \vz_\text{AR}^i)$ are then fed to $\vpsi_f^i(\cdot)$, which outputs the FiLM parameters for layer $i$. Green arrows correspond to propagation of auto-regressive representations. Note that the auto-regressive component $\vz_\text{AR}^i$ is computed by processing the \textit{adapted} activations $\{f_\theta^i(\vx ; \vpsi_f^{\tau}) \}$ of the previous convolutional block.} 
    \label{fig:auto_regressive_feature_adaptation}
\end{figure}

\begin{figure}
    \centering
    \includegraphics[width=0.5\textwidth]{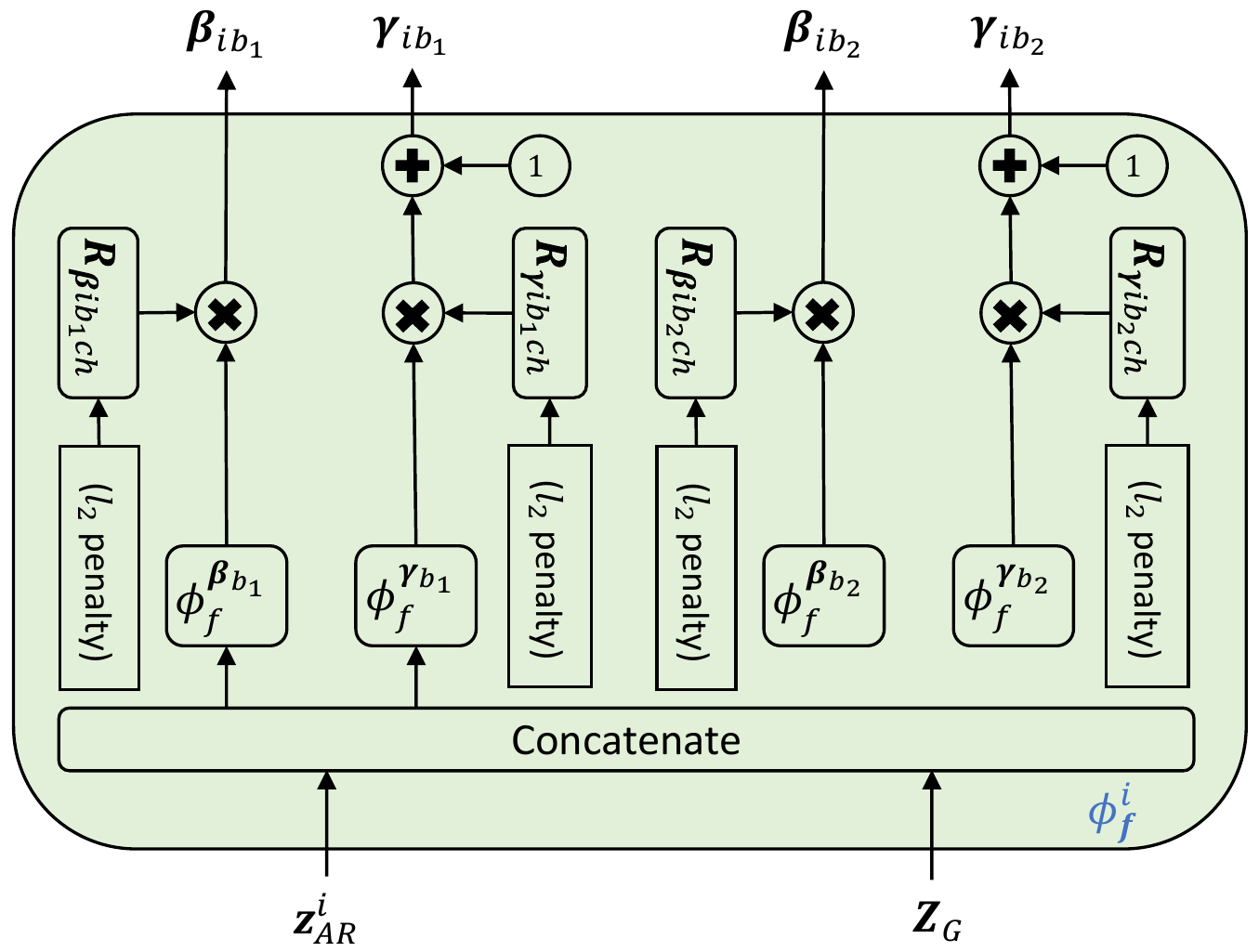}
    \caption{Adaptation network $\vphi_f$. $\bm{R}_{\gamma{i}{b_j}ch}$ and $\bm{R}_{\beta{i}{b_j}ch}$ denote a vector of regularization weights that are learned with an $l_2$ penalty.}
    \label{fig:film_generators}
\end{figure}

To adapt the $l^{\text{th}}$ layer in the feature extractor, it is useful for the system to have access to the representation of task-relevant inputs from layer $l-1$. While $\vz_G$ could in principle encode how layer $l-1$ has adapted, we opt to provide this information directly to the adaptation network adapting layer $l$ by passing the adapted activations from layer $l-1$. The auto-regressive component $\vz_\text{AR}^i$ is computed by processing the \textit{adapted} activations of the previous convolutional block with a layer-specific set encoder (except for the first residual block, whose auto-regressive component is given by the \textit{un-adapted} initial pre-processing stage in the ResNet). Both the global and all layer-specific set-encoders are implemented as permutation invariant functions \citep{zaheer2017deep,qi2017pointnet} (see \cref{app:network_architecture_details} for details). The full parameterization is illustrated in \cref{fig:auto_regressive_feature_adaptation}, and the architecture of $\vpsi^i_f(\cdot)$ networks is illustrated in \cref{fig:film_generators}.

\section{Model Training}
\label{sec:training}
The previous section has specified the model (see \cref{fig:architexture} for a schematic). We now describe how to train the global classifier parameters $\vtheta$ and the adaptation network parameters $\vphi = \{\vphi_f, \vphi_w \}$. 

\paragraph{Training the global classifier parameters $\vtheta$.} A natural approach to training the model (originally employed by CNPs \citep{garnelo2018conditional}) would be to maximize the likelihood of the training data jointly over $\vtheta$ and $\vphi$. However, experiments (detailed in \cref{app:training_experiments}) showed that it is crucially important to adopt a two stage process instead. In the first stage, $\vtheta$ are trained on a large dataset (e.g., the training set of ImageNet \citep{russakovsky2015imagenet, triantafillou2019meta}) in a full-way classification procedure, mirroring standard pre-training. Second, $\vtheta$ are fixed and $\vphi$ are trained using episodic training over all meta-training datasets in the multi-task setting. We hypothesize that two-stage training is important for two reasons:
\begin{inlinelist}
\item during the second stage, $\vphi_f$ are trained to adapt $f_\vtheta(\cdot)$ to tasks $\tau$ by outputting $\vpsi^\tau_f$. As $\vtheta$ has far more capacity than $\vpsi^\tau_f$, if they are trained in the context of all tasks, there is no need for $\vpsi^\tau_f$ to adapt the feature extractor, resulting in little-to-no training signal for $\vphi_f$ and poor generalisation.
\item Allowing $\vtheta$ to adapt during the second phase violates the principle of ``train as you test'', i.e., when test tasks are encountered, $\vtheta$ will be fixed, so it is important to simulate this scenario during training.
\end{inlinelist}
Finally, fixing $\vtheta$ during meta-training is desireable as it results in a dramatic decrease in training time.
\label{sec:hypernet_training}

\paragraph{Training the adaptation network parameters $\vphi$.} Following the work of \citet{garnelo2018conditional}, we train $\vphi$ with maximum likelihood. An unbiased stochastic estimator of the log-likelihood is:
\begin{equation}
\label{eqn:stochastic_bdt_ml}
\hat{\mathcal{L}} \left( \vphi \right)  = \frac{1}{MT} \sum\limits_{m, \tau} \log p\left( \vy^{\ast\tau}_m | \vx^{\ast\tau}_m, \vpsi_\vphi \left( D^\tau \right), \vtheta \right),
\end{equation}
where $\{\vy^{\ast\tau}_m, \vx^{\ast\tau}_m, D^\tau\} \sim \hat{P}$, with $\hat{P}$ representing the data distribution (e.g., sampling tasks and splitting them into disjoint context ($D^\tau$) and target data $\{ (\vx_m^{\ast\tau}, \vy_m^{\ast\tau}) \}_{m=1}^{M_t}$). Maximum likelihood training therefore naturally uses episodic context / target splits often used in meta-learning. In our experiments we use the protocol defined by \citet{triantafillou2019meta} and \textsc{Meta-Dataset} for this sampling procedure. \cref{alg:stochastic_estimator} details computation of the stochastic estimator for a single task.

\ifstandalone
\newpage
\bibliography{bibliography}
\bibliographystyle{abbrvnat}
\fi

\section{Related Work}
\label{sec:related}

\begin{figure}[t]
\floatbox[{\capbeside\thisfloatsetup{capbesideposition={right,top},capbesidewidth=0.48\textwidth}}]{figure}[\FBwidth]
{\caption{Model design space. The $y$-axis represents the number of task-specific parameters $|\vpsi^\tau|$. Increasing $|\vpsi^\tau|$ increases model flexibility, but also the propensity to over-fit. The $x$-axis represents the complexity of the mechanism used to adapt the task-specific parameters to training data $\vpsi(D^\tau)$.  On the right are \textit{amortized} approaches (i.e.~using fixed functions). On the left is gradient-based adaptation. Mixed approaches lie between. Computational efficiency increases to the right. Flexibility increases to the left, but with it over-fitting and need for hand tuning.}
\label{fig:related_plot}}
{\includegraphics[width=0.47\textwidth]{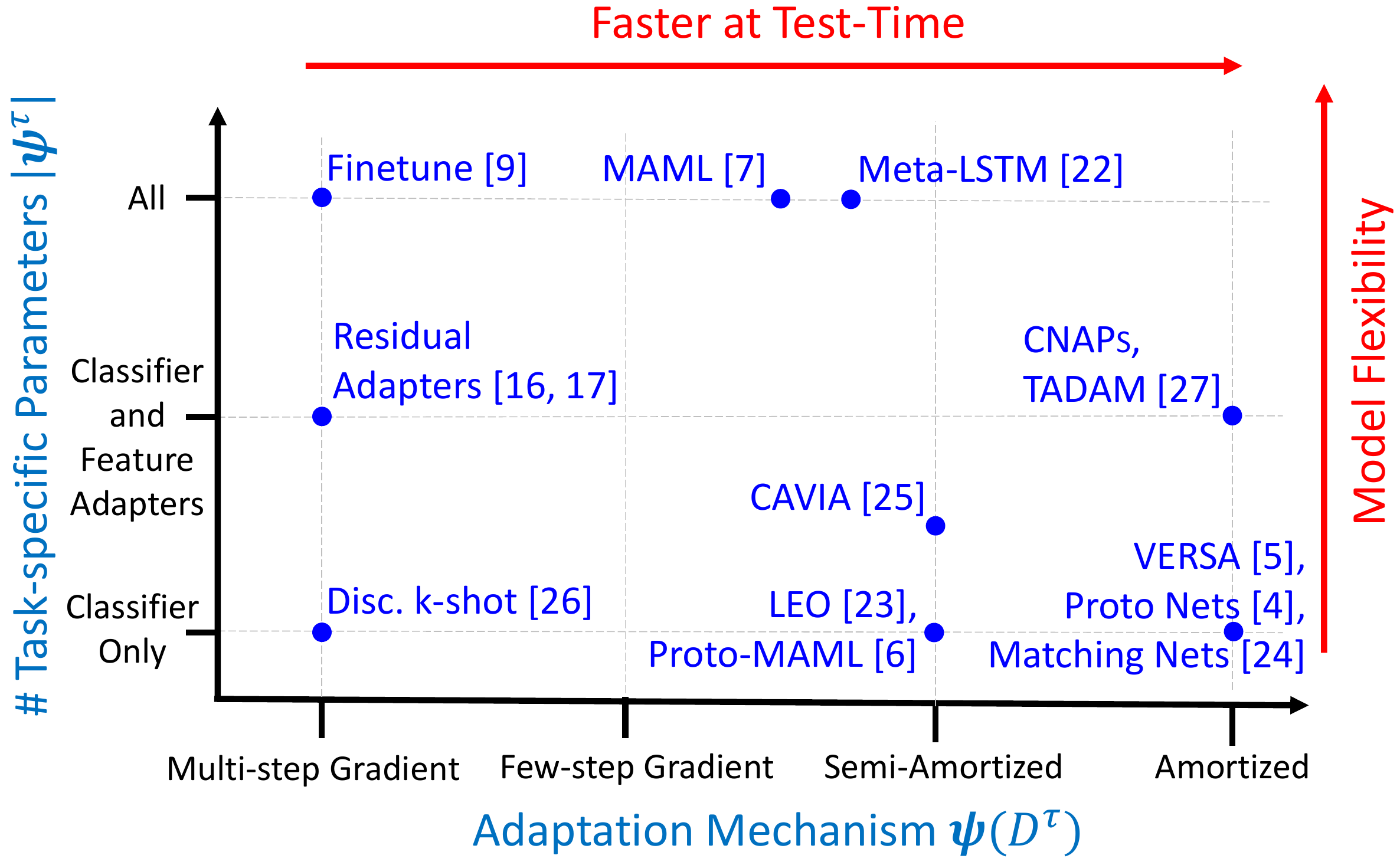}}
\end{figure}

Our work frames multi-task classification as directly modelling the predictive distribution $p(\vy^\ast | \vx^\ast, \vpsi(D^\tau))$. The perspective allows previous work \citep{finn2017model,gordon2018meta,perez2018film,ravi2016optimization,rebuffi2017learning,rebuffi2018efficient,rusu2018meta,snell2017prototypical,triantafillou2019meta,vinyals2016matching,yosinski2014transferable,zintgraf2018caml,bauer2017discriminative} to be organised in terms of i) the choice of the parameterization of the classifier (and in particular the nature of the local parameters), and ii) the function used to compute the local parameters from the training data. This space is illustrated in \cref{fig:related_plot}, and further elaborated upon in \cref{app:related_work}.

One of the inspirations for our work is conditional neural processes (CNPs) \citep{garnelo2018conditional}. CNPs directly model the predictive distribution $p(\vy^\ast | \vx^\ast, \vpsi(D^\tau))$ and train the parameters using maximum likelihood. Whereas previous work on CNPs has focused on homogeneous regression and classification datasets and fairly simple models, here we study multiple heterogeneous classification datasets and use a more complex model to handle this scenario. In particular, whereas the original CNP approach to classification required pre-specifying the number of classes in advance, \cnaps{} handles varying way classification tasks, which is required for e.g.~the meta-dataset benchmark. Further, \cnaps{} employs a parameter-sharing hierarchy that parameterizes the feature extractor. This contrasts to the original CNP approach that shared all parameters across tasks, and use latent inputs to the decoder to adapt to new tasks. Finally, \cnaps{} employs a meta-training procedure geared towards \textit{learning to adapt} to diverse tasks. Similarly, our work can be viewed as a deterministic limit of ML-PIP \citep{gordon2018meta} which employs a distributional treatment of the local-parameters $\vpsi$.

A model with design choices closely related to \cnaps{} is TADAM \citep{oreshkin2018tadam}. TADAM employs a similar set of local parameters, allowing for adaptation of both the feature extractor and classification layer. However, it uses a far simpler adaptation network (lacking auto-regressive structure) and an expensive and ad-hoc training procedure. Moreover, TADAM was applied to simple few-shot learning benchmarks (e.g. CIFAR100 and mini-ImageNet) and sees little gain from feature extractor adaptation. In contrast, we see a large benefit from adapting the feature extractor. This may in part reflect the differences in the two models, but we observe that feature extractor adaptation has the largest impact when used to adapt to \textit{different datasets} and that two stage training is required to see this. Further differences are our usage of the CNP framework and the  flexible deployment of \cnaps{} to continual learning and active learning (see \cref{sec:experiments}).

\ifstandalone
\newpage
\bibliography{bibliography}
\bibliographystyle{abbrvnat}
\fi

\section{Experiments and Results}
\label{sec:experiments}
The experiments target three key questions:
\begin{inlinelist}
    \item Can \cnaps{} improve performance in multi-task few-shot learning?
    \item Does the use of an adaptation network benefit computational-efficiency and data-efficiency? 
    \item Can \cnaps{} be deployed directly to complex learning scenarios like continual learning and active learning?
\end{inlinelist}
The experiments use the following modelling choices (see \cref{app:network_architecture_details} for full details). While \cnaps{} can utilize any feature extractor, a ResNet18 \citep{he2016deep} is used throughout to enable fair comparison with \citet{triantafillou2019meta}. 
To ensure that each task is handled independently, batch normalization statistics \citep{ioffe2015batch} are learned (and fixed) during the pre-training phase for $\vtheta$. Actual batch statistics of the test data are never used during meta-training or testing. 

\paragraph{Few Shot Classification.} The first experiment tackles a demanding few-shot classification challenge called \textsc{Meta-Dataset} \citep{triantafillou2019meta}. \textsc{Meta-Dataset} is composed of ten (eight train, two test) image classification datasets. The challenge constructs few-shot learning tasks by drawing from the following distribution. First, one of the datasets is sampled uniformly; second, the ``way'' and ``shot'' are sampled randomly according to a fixed procedure; third, the classes and context / target instances are sampled. Where a hierarchical structure exists in the data (ILSVRC or \textsc{Omniglot}), task-sampling respects the hierarchy. In the meta-test phase, the identity of the original dataset is not revealed and the tasks must be treated independently (i.e.~no information can be transferred between them). Notably, the meta-training set comprises a disjoint and dissimilar set of classes from those used for meta-test. Full details are available in \cref{app:meta_dataset_procedure} and  \citep{triantafillou2019meta}.

\citet{triantafillou2019meta} consider two stage training: an initial stage that trains a feature extractor in a standard classification setting, and a meta-training stage of all parameters in an episodic regime. For the meta-training stage, they consider two settings: meta-training only on the \textsc{Meta-Dataset} version of ILSVRC, and on all meta-training data. We focus on the latter as \cnaps{} rely on training data from a variety of training tasks to learn to adapt, but provide results for the former in \cref{app:meta_dataset_imagenet}. We pre-train $\vtheta$ on the meta-training set of the \textsc{Meta-Dataset} version of ILSVRC, and meta-train $\vphi$ in an episodic fashion using all meta-training data. We compare \cnaps{} to models considered by \citet{triantafillou2019meta}, including their proposed method (Proto-MAML) in \cref{table:meta_dataset_results_all}. We meta-test \cnaps{} on three additional held-out datasets: MNIST \citep{lecun2010mnist}, CIFAR10 \citep{krizhevsky2009learning}, and CIFAR100 \citep{krizhevsky2009learning}. As an ablation study, we compare a version of \cnaps{} that does not make use of the auto-regressive component $\vz_{AR}$, and a version that uses no feature extractor adaptation. In our analysis of \cref{table:meta_dataset_results_all}, we distinguish between two types of generalization: \begin{inlinelist}
    \item unseen tasks (classes) in meta-training datasets, and
    \item unseen datasets.
\end{inlinelist}



\paragraph{Unseen tasks:} \cnaps{} achieve significant improvements over existing methods on seven of the eight datasets. The exception is the \textsc{Textures} dataset, which has only seven test classes and accuracy is highly sensitive to the train / validation / test class split. The ablation study demonstrates that removing $\vz_{\text{AR}}$ from the feature extractor adaptation degrades accuracy in most cases, and that removing all feature extractor adaptation results in drastic reductions in accuracy.

\paragraph{Unseen datasets:} \cnaps-models outperform all competitive models with the exception of \textsc{Finetune} on the \textsc{Traffic Signs} dataset. Removing $\vz_{\text{AR}}$ from the feature extractor decreases accuracy and removing the feature extractor adaptation entirely significantly impairs performance. The degradation is particularly pronounced when the held out dataset differs substantially from the dataset used to pretrain $\vtheta$, e.g.~for MNIST.

 Note that the superior results when using the auto-regressive component can not be attributed to increased network capacity alone. In \cref{app:parallel_residual_adapter_comparison} we demonstrate that \cnaps{} yields superior classification accuracy when compared to parallel residual adapters \citep{rebuffi2018efficient} even though \cnaps{} requires significantly less network capacity in order to adapt the feature extractor to a given task.

\begin{table}[t]
    \caption{Few-shot classification results on \textsc{Meta-Dataset} \citep{triantafillou2019meta} using models trained on all training datasets. All figures are percentages and the $\pm$ sign indicates the 95\% confidence interval over tasks. Bold text indicates the scores within the confidence interval of the highest score. Tasks from datasets below the dashed line were not used for training. Competing methods' results from \citep{triantafillou2019meta}.} 
    \label{table:meta_dataset_results_all}
	\centering
	\tiny
	\begin{tabular}{lccccccccc}
	\toprule
	Dataset                                   & Finetune                &  MatchingNet    &  ProtoNet      & fo-MAML         & Proto-MAML        &  \multicolumn{1}{c}{\begin{tabular}[c]{@{}c@{}} \cnaps{} \\ (no $\vpsi_f$)\end{tabular}} & \multicolumn{1}{c}{\begin{tabular}[c]{@{}c@{}} \cnaps{} \\ (no $\vz_{AR}$)\end{tabular}}          & \cnaps{}          \\
    \midrule
	ILSVRC \citep{russakovsky2015imagenet}    & 43.1 $\pm$ 1.1          & 36.1 $\pm$ 1.0  & 44.5 $\pm$ 1.1 & 32.4 $\pm$ 1.0  & 47.9 $\pm$ 1.1           & 43.8 $\pm$ 1.0            & \textbf{51.3 $\pm$ 1.0}   & \textbf{52.3 $\pm$ 1.0} \\
	Omniglot \citep{lake2011one}              & 71.1 $\pm$ 1.4          & 78.3 $\pm$ 1.0  & 79.6 $\pm$ 1.1 & 71.9 $\pm$ 1.2  & 82.9 $\pm$ 0.9           & 60.1 $\pm$ 1.3            & \textbf{88.0 $\pm$ 0.7}   & \textbf{88.4 $\pm$ 0.7} \\
    Aircraft \citep{maji2013fine}             & 72.0 $\pm$ 1.1          & 69.2 $\pm$ 1.0  & 71.1 $\pm$ 0.9 & 52.8 $\pm$ 0.9  & 74.2 $\pm$ 0.8           & 53.0 $\pm$ 0.9            & 76.8 $\pm$ 0.8            & \textbf{80.5 $\pm$ 0.6} \\
    Birds \citep{wah2011caltech}              & 59.8 $\pm$ 1.2          & 56.4 $\pm$ 1.0  & 67.0 $\pm$ 1.0 & 47.2 $\pm$ 1.1  & 70.0 $\pm$ 1.0           & 55.7 $\pm$ 1.0            & \textbf{71.4 $\pm$ 0.9}   & \textbf{72.2 $\pm$ 0.9} \\
    Textures \citep{cimpoi2014describing}     & \textbf{69.1 $\pm$ 0.9} & 61.8 $\pm$ 0.7  & 65.2 $\pm$ 0.8 & 56.7 $\pm$ 0.7  & 67.9 $\pm$ 0.8           & 60.5 $\pm$ 0.8            & 62.5 $\pm$ 0.7            & 58.3 $\pm$ 0.7 \\
    Quick Draw \citep{ha2017neural}           & 47.0 $\pm$ 1.2          & 60.8 $\pm$ 1.0  & 64.9 $\pm$ 0.9 & 50.5 $\pm$ 1.2  & 66.6 $\pm$ 0.9           & 58.1 $\pm$ 1.0            & \textbf{71.9 $\pm$ 0.8}   & \textbf{72.5 $\pm$ 0.8} \\
    Fungi \citep{Schroeder2018FGVCx}          & 38.2 $\pm$ 1.0          & 33.7 $\pm$ 1.0  & 40.3 $\pm$ 1.1 & 21.0 $\pm$ 1.0  & 42.0 $\pm$ 1.1           & 28.6 $\pm$ 0.9            & \textbf{46.0 $\pm$ 1.1}   & \textbf{47.4 $\pm$ 1.0} \\
    VGG Flower \citep{nilsback2008automated}  & 85.3 $\pm$ 0.7          & 81.9 $\pm$ 0.7  & 86.9 $\pm$ 0.7 & 70.9 $\pm$ 1.0  & \textbf{88.5 $\pm$ 0.7}  & 75.3 $\pm$ 0.7            & \textbf{89.2 $\pm$ 0.5}   & 86.0 $\pm$ 0.5 \\
	\hdashline
    Traffic Signs \citep{houben2013detection} & \textbf{66.7 $\pm$ 1.2} & 55.6 $\pm$ 1.1  & 46.5 $\pm$ 1.0 & 34.2 $\pm$ 1.3  & 52.3 $\pm$ 1.1           & 55.0 $\pm$ 0.9            & 60.1 $\pm$ 0.9            & 60.2 $\pm$ 0.9 \\ 
    MSCOCO \citep{lin2014microsoft}           & 35.2 $\pm$ 1.1          & 28.8 $\pm$ 1.0  & 39.9 $\pm$ 1.1 & 24.1 $\pm$ 1.1  & \textbf{41.3 $\pm$ 1.0}  & \textbf{41.2 $\pm$ 1.0}   & \textbf{42.0 $\pm$ 1.0}   & \textbf{42.6 $\pm$ 1.1} \\
    MNIST \citep{lecun2010mnist}              &                         &                 &  & &                                                        & 76.0 $\pm$ 0.8            & 88.6 $\pm$ 0.5            & \textbf{92.7 $\pm$ 0.4} \\
    CIFAR10 \citep{krizhevsky2009learning}    &                         &                 &  & &                                                        & \textbf{61.5 $\pm$ 0.7}   & \textbf{60.0 $\pm$ 0.8}   & \textbf{61.5 $\pm$ 0.7} \\ 
    CIFAR100 \citep{krizhevsky2009learning}   &                         &                 &  & &                                                        & 44.8 $\pm$ 1.0            & \textbf{48.1 $\pm$ 1.0}   & \textbf{50.1 $\pm$ 1.0} \\
    \bottomrule
\end{tabular}
\end{table}

\begin{figure}[t]
\centering
\includegraphics[width=1\textwidth]{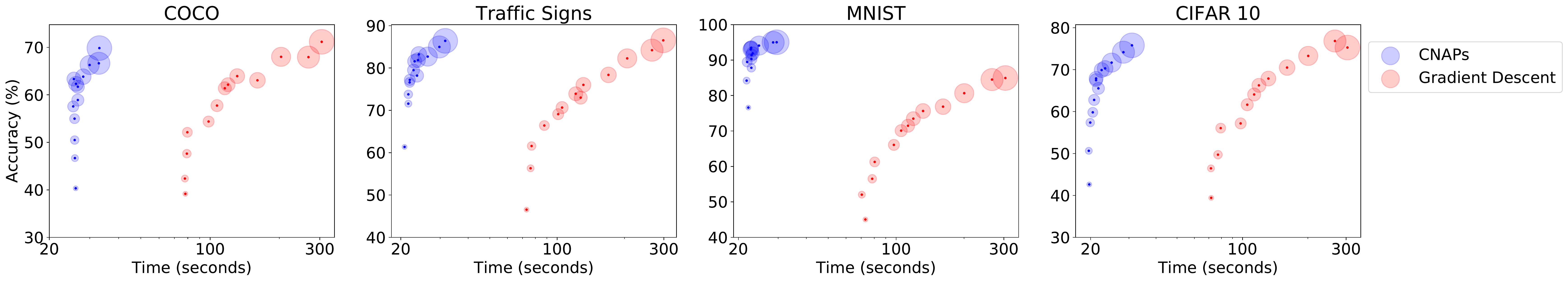}
\caption{Comparing \cnaps{} to gradient based feature extractor adaptation: accuracy on 5-way classification tasks from withheld datasets as a function of processing time. Dot size reflects shot number ($1$ to $25$ shots).}
\label{fig:amortize_versus_gradient_time_scatter}
\end{figure}

\begin{figure}
\hspace{-1.5cm}
\begin{minipage}{0.5\textwidth}
\centering
\includegraphics[width=1\linewidth]{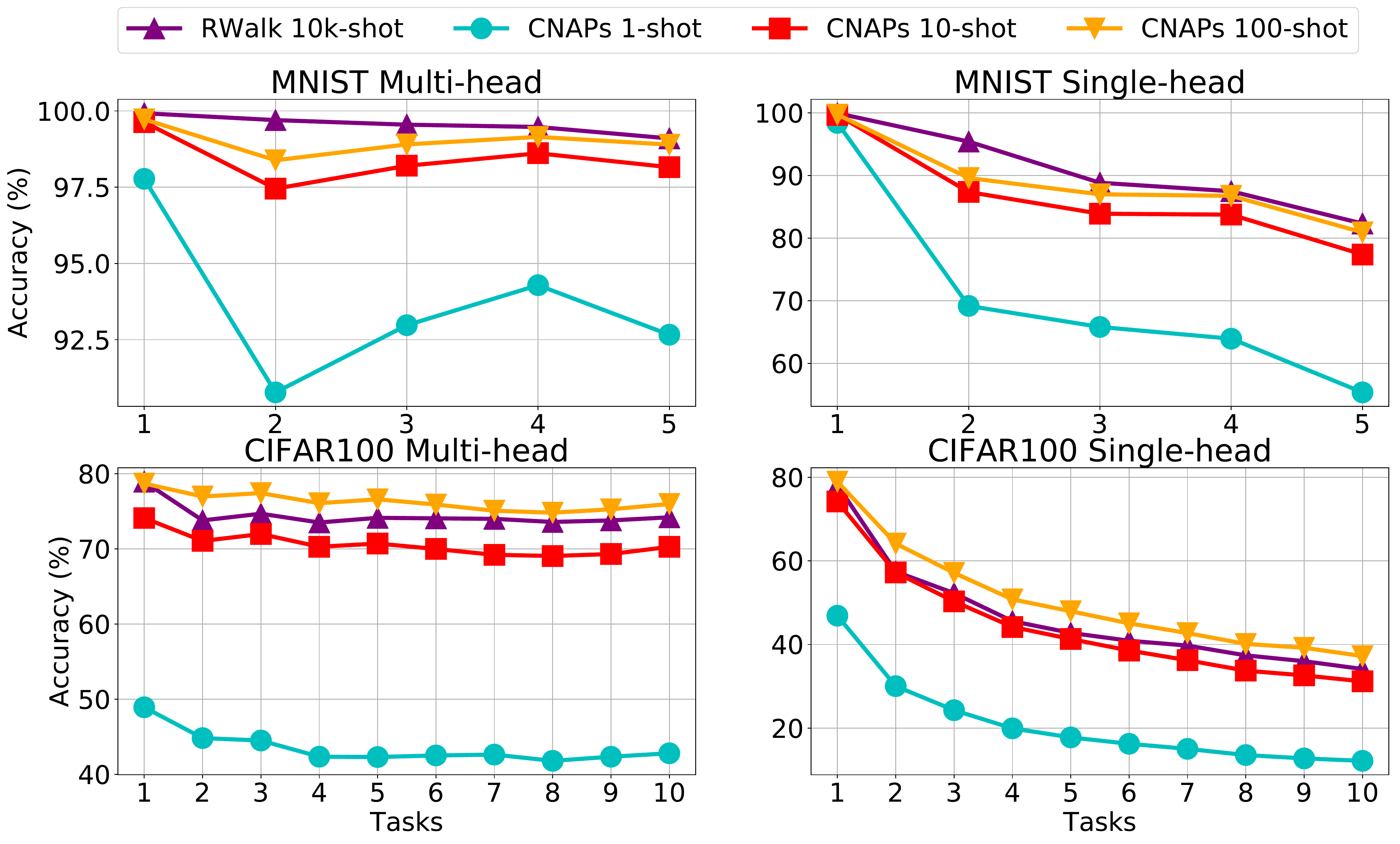}
\end{minipage}%
\hspace{0.2cm}
\begin{minipage}{0.4\textwidth}
\small
\centering
\captionsetup{type=table} 
	\begin{tabular}{lcccccc}
        &\multicolumn{2}{c}{\textit{MNIST}} & \multicolumn{2}{c}{\textit{CIFAR100}} \\ 
        Method                                & Multi & Single &  Multi & Single \\
        \midrule
        SI \citep{zenke2017continual}                          & 99.3      & 57.6               & 73.2                  & 22.8               \vspace{1mm}\\
        EWC \citep{kirkpatrick2017overcoming}                  & 99.3      & 55.8               & 72.8                  & 23.1               \vspace{1mm}\\
        {VCL \citep{nguyen2017variational}}     & 98.5      & \multirow{2}{*}{-} & \multirow{2}{*}{-}    & \multirow{2}{*}{-} \\
                                                               & $\pm$ 0.4 &                    &                       &                    \vspace{1mm}\\
        RWalk \citep{chaudhry2018riemannian}                   & 99.3      & 82.5               & 74.2                  & 34.0               \vspace{1mm}\\
        {\cnaps}                                & 98.9      & 80.9               & 76.0                  & 37.2               \\
                                                               & $\pm$ 0.2 & $\pm$ 0.9          & $\pm$ 0.5             & $\pm$ 0.6          \vspace{1mm}\\

        \bottomrule
\end{tabular}
\end{minipage}
\caption{Continual learning classification results on Split MNIST and Split CIFAR100 using a model trained on all training datasets. (Left) The plots show accumulated accuracy averaged over 30 runs for both single- and multi-head scenarios. (Right) Average accuracy at final task computed over 30 experiments (all figures are percentages). Errors are one standard deviation. Additional results from \citep{chaudhry2018riemannian,swaroop2019improving}.}
\label{fig:continual_learning_results}
\end{figure}

\paragraph{Additional results:} Results when meta-training only on the \textsc{Meta-Dataset} version of ILSVRC are given in \cref{table:meta_dataset_results_imagenet}.
In \cref{app:film_visualization}, we visualize the task encodings and parameters, demonstrating that the model is able to learn meaningful task and dataset level representations and parameterizations. The results support the hypothesis that learning to adapt key parts of the network is more robust and achieves significantly better performance than existing approaches. 
\paragraph{FiLM Parameter Learning Performance: Speed-Accuracy Trade-off.}
%
\cnaps{} generate FiLM layer parameters for each task $\tau$ at test time using the adaptation network $\vpsi_f(D^{\tau})$. It is also possible to learn the FiLM parameters via gradient descent (see \citep{rebuffi2017learning, rebuffi2018efficient}). Here we compare \cnaps{} to this approach. \cref{fig:amortize_versus_gradient_time_scatter} shows plots of 5-way classification accuracy versus time for four held out data sets as the number of shots was varied. For gradient descent, we used a fixed learning rate of 0.001 and took 25 steps for each point. The overall time required to produce the plot was 1274 and 7214 seconds for \cnaps{} and gradient approaches, respectively, on a NVIDIA Tesla P100-PCIE-16GB GPU. \cnaps{} is at least 5 times faster at test time than gradient-based optimization requiring only a single forward pass through the network while gradient based approaches require multiple forward and backward passes. Further, the accuracy achieved with adaptation networks is significantly higher for fewer shots as it protects against over-fitting. For large numbers of shots, gradient descent catches up, albeit slowly. 
   \paragraph{Complex Learning Scenarios: Continual Learning.} In continual learning \citep{ring1997child} new tasks appear over time and existing tasks may change. The goal is to adapt accordingly, but without retaining old data which is challenging for artificial systems. To demonstrate the the versatility \cnaps{} we show that, although it has not been explicitly trained for continual learning, we are able to apply the same model trained for the few-shot classification experiments (without the auto-regressive component) to standard continual learning benchmarks on held out datasets: Split MNIST \citep{zenke2017continual} and Split CIFAR100 \citep{chaudhry2018riemannian}. We modify the model to compute running averages for the representations of both $\vpsi^\tau_w$ and $\vpsi^\tau_f$ (see \cref{app:additional_continual_learning_details} for further details), in this way it performs incremental updates using the new data and the old model, and does not need to access old data. \cref{fig:continual_learning_results} (left) shows the accumulated multi-  and single-head \citep{chaudhry2018riemannian} test accuracy averaged over 30 runs (further results and more detailed figures are in \cref{app:additional_continual_learning_results}). \cref{fig:continual_learning_results} (right) shows average results at the final task comparing to SI \citep{zenke2017continual}, EWC \citep{kirkpatrick2017overcoming}, VCL \citep{nguyen2017variational}, and Riemannian Walk \citep{chaudhry2018riemannian}. 

\cref{fig:continual_learning_results} demonstrates that \cnaps{} naturally resists catastrophic forgetting \citep{kirkpatrick2017overcoming} and compares favourably to competing methods, despite the fact that it was not exposed to these datasets during training, observes orders of magnitude fewer examples, and was not trained explicitly to perform continual learning. \cnaps{} performs similarly to, or better than, the state-of-the-art Riemannian Walk method which departs from the pure continual learning setting by maintaining a small number of training samples across tasks. Conversely, \cnaps{} has the advantage of being exposed to a larger range of datasets and can therefore leverage task transfer. We emphasize that this is not meant to be an ``apples-to-apples'' comparison, but rather, the goal is to demonstrate the out-of-the-box versatility and strong performance of \cnaps{} in new domains and learning scenarios.
\begin{figure}[t]
\centering
\includegraphics[width=1.0\textwidth]{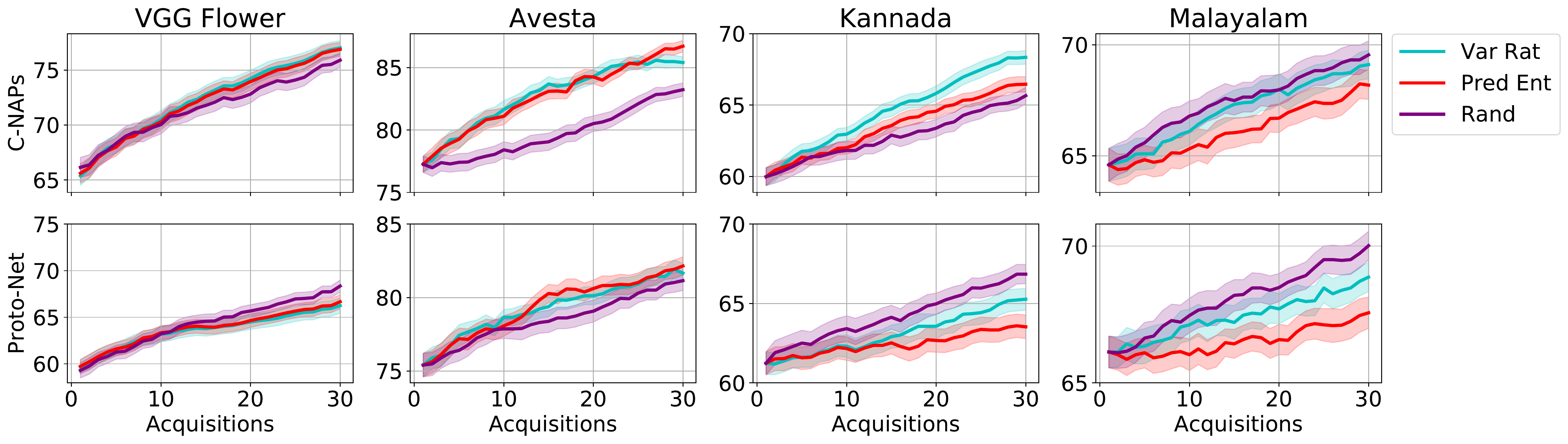}
\caption{Accuracy vs active learning iterations for held-out classes / languages. (Top) \cnaps{} and (bottom) prototypical networks. Error shading is one standard error. \cnaps{} achieves better accuracy than prototypical networks and improvements over random acquisition, whereas prototypical networks do not.}
\label{fig:active_learning}\
\end{figure}

\textbf{Complex Learning Scenarios: Active Learning}. Active learning \cite{cohn1996active, settles2012active} requires accurate data-efficient learning that returns well-calibrated uncertainty estimates. \cref{fig:active_learning} compares the performance of \cnaps{} and prototypical networks using two standard active learning acquisition functions (variation ratios and predictive entropy \citep{cohn1996active}) against random acquisition on the \textsc{Flowers} dataset and three representative held-out languages from \textsc{Omniglot} (performance on all languages is presented in \cref{app:additional_active_learning_results}). \cref{fig:active_learning,app:additional_active_learning_results} show that \cnaps{} achieves higher accuracy on average than prototypical networks. Moreover, \cnaps{} achieves significant improvements over random acquisition, whereas prototypical networks do not. These tests indicates that \cnaps{} is more accurate and suggest that \cnaps{} has better calibrated uncertainty estimates than prototypical networks.

\ifstandalone
\newpage
\bibliography{bibliography}
\bibliographystyle{abbrvnat}
\fi

\section{Conclusions}
\label{sec:conclusions}

This paper has introduced \cnaps{}, an automatic, fast and flexible modelling approach for multi-task classification. We have demonstrated that \cnaps{}   achieve state-of-the-art performance on the \textsc{Meta-Dataset} challenge, and can be deployed ``out-of-the-box'' to diverse learning scenarios such as continual and active learning where they are competitive with the state-of-the-art. Future avenues of research are to consider the exploration of the design space by introducing gradients and function approximation to the adaptation mechanisms, as well as generalizing the approach to distributional extensions of \cnaps{} \citep{garnelo2018neural,kim2018attentive}.


\ifstandalone
\newpage
\bibliography{bibliography}
\bibliographystyle{abbrvnat}
\fi

\section*{Acknowledgments}
The authors would like to thank Ambrish Rawat for helpful discussions and David Duvenaud, Wessel Bruinsma, Will Tebbutt Adri\`{a} Garriga Alonso, Eric Nalisnick,  and Lyndon White for the insightful comments and feedback. Richard E. Turner is supported by Google, Amazon, Improbable and EPSRC grants EP/M0269571 and EP/L000776/1.

\bibliography{main}
\bibliographystyle{unsrtnat}


\clearpage
\newpage

\newpage
\appendix
\title{Supplementary Material for Fast and Flexible Multi-Task Classification Using Conditional Neural Adaptive Processes}
\author{
 James Requeima\footnotemark[1] \\
 University of Cambridge \\
 Invenia Labs \\
 \texttt{jrr41@cam.ac.uk} \\
 \And
 Jonathan Gordon\footnotemark[1] \\
 University of Cambridge \\
 \texttt{jg801@cam.ac.uk} \\
 \And
 John Bronskill\footnotemark[1] \\
 University of Cambridge \\
 \texttt{jfb54@cam.ac.uk} \\
 \And
 Sebastian Nowozin \\
 Google Research Berlin \\
 \texttt{nowozin@google.com}\\
 \And
 Richard E.~Turner \\
 University of Cambridge \\
 Microsoft Research \\
 \texttt{ret26@cam.ac.uk}
}


\section{Algorithm for Constructing Stochastic Estimator}
\label{app:stochastic_estimator}

An algorithm for constructing the stochastic training objective  $\hat{\mathcal{L}}(\vphi;\tau)$ for a single task $\tau$ is given in \cref{alg:stochastic_estimator}. $\text{CAT}(\cdot ; \vpi)$ denotes a the likelihood of a categorical distribution with parameter vector $\vpi$. This algorithm can be used on a batch of tasks to construct an unbiased estimator for the auto-regressive likelihood of the task outputs.

\begin{algorithm}[h]
\caption{Stochastic Objective Estimator for Meta-Training.}\label{alg:stochastic_estimator}
\begin{algorithmic}[1]
\Procedure{Meta-Training}{$\{\vx^\ast_m, \vy^\ast_m\}_{m=1}^M, D^\tau, \vtheta, \vphi$}
\State $\vpsi^\tau_f \leftarrow \vpsi_f(\{ f_\vtheta(\vx_n) | \vx \in D^\tau\}; \vphi_f)$
\State $\vpsi^\tau_c \leftarrow \vpsi_w(\{ f_\vtheta(\vx_n; \vpsi_f) | \vx \in D^\tau, \vy_n=c\}; \vphi_w) \quad \forall c \in C^\tau$
\For{\texttt{$m \in 1,...,M$}}
    \State $\vpi_m \leftarrow f_\vtheta (\vx^\ast_m; \vpsi^\tau_f)^T \vpsi^\tau_w$ 
    \State $\log p(\vy_m^\ast | \vpi_m ) \leftarrow \log \text{CAT}(\vy^\ast_m ; \vpi_m)$ 
\EndFor
\State \textbf{return} $\hat{\mathcal{L}}(\vphi; \tau) \leftarrow \frac{1}{M}\sum\limits_M \log p(\vy_m^\ast | \vpi_m)$
\EndProcedure
\end{algorithmic}
\end{algorithm}

\section{Additional Related Work Details}
\label{app:related_work}

\paragraph{The choice of task-specific parameters $\vpsi^{\tau}$.} Clearly, any approach to multi-task classification must adapt, at the very least, the top-level classifier layer of the model. A number of successful models have proposed doing just this with e.g., neighbourhood-based approaches \citep{snell2017prototypical}, variational inference \citep{bauer2017discriminative}, or inference networks \citep{gordon2018meta}. On the other end of the spectrum are models that adapt \textit{all} the parameters of the classifier, e.g., \citep{finn2017model,nichol2018reptile,kim2018bayesian}. The trade-off here is clear: as more parameters are adapted, the resulting model is more flexible, but also slow and prone to over-fitting. For this reason we modulate a small portion of the network parameters, following recent work on multi-task learning \citep{rebuffi2017learning, rebuffi2018efficient, perez2018film}.

We argue that just adapting the linear classification layer is sufficient when the task distribution is not diverse, as in the standard benchmarks used for few-shot classification (OMNIGLOT \citep{lake2011one} and \textit{mini}-imageNet \citep{ravi2016optimization}). However, when faced with a diverse set of tasks, such as that introduced recently by \citet{triantafillou2019meta}, it is important to adapt the feature extractor on a per-task basis as well. 

\paragraph{The adaptation mechanism $\vpsi_{\phi} \left( D^\tau \right)$.} Adaptation varies in the literature from performing full gradient descent learning with $D^\tau$ \cite{yosinski2014transferable} to relying on simple operations such as taking the mean of class-specific feature representations \citep{snell2017prototypical,vinyals2016matching}. Recent work has focused on reducing the number of required gradient steps by learning a global initialization \citep{finn2017model,nichol2018reptile} or additional parameters of the optimization procedure \citep{ravi2016optimization}. Gradient-based procedures have the benefit of being flexible, but are computationally demanding, and prone to over-fitting in the low-data regime. Another line of work has focused on learning neural networks to output the values of $\vpsi$, which we denote \textit{amortization} \citep{gordon2018meta}. Amortization greatly reduces the cost of adaptation and enables sharing of global parameters, but may suffer from the amortization gap \citep{cremer2018inference} (i.e., underfitting), particularly in the large data regime. Recent work has proposed using semi-amortized inference \citep{triantafillou2019meta,rusu2018meta}, but have done so while only adapting the classification layer parameters.

\section{Experimentation Details}
\label{app: experimentation_details}
All experiments were implemented in PyTorch \citep{paszke2017automatic} and executed either on NVIDIA Tesla P100-PCIE-16GB or Tesla V100-SXM2-16GB GPUs. The full \cnaps{} model runs in a distributed fashion across 2 GPUs and takes approximately one and a half days to complete episodic training and testing.

\subsection{\textsc{Meta-Dataset} Training and Evaluation Procedure}
\label{app:meta_dataset_procedure}

\subsubsection{Feature Extractor Weights \texorpdfstring{$\vtheta$}{TEXT} Pretraining}
\label{app:feature_extractor_pretraining}
We first reduce the size of the images in the ImageNet ILSVRC-2012 dataset \citep{russakovsky2015imagenet} to 84 $\times$ 84 pixels. Some images in the ImageNet ILSVRC-2012 dataset are duplicates of images in other datasets included in \textsc{Meta-Dataset}, so these were removed. We then split the 1000 training classes of the ImageNet ILSVRC-2012 dataset into training, validation, and test sets according to the criteria detailed in \citep{triantafillou2019meta}. The test set consists of the 130 leaf-node subclasses of the ``device" synset node, the validation set consists of the the 158 leaf-node subclasses of the ``carnivore" synset node, and the training set consists of the remaining 712 leaf-node classes. We then pretrain a feature extractor with parameters $\theta$ based on a modified ResNet-18 \citep{he2016deep} architecture on the above 712 training classes. The ResNet-18 architecture is detailed in \cref{table:pre_trained_resnet_architecture}. Compared to a standard ResNet-18, we reduced the initial convolution kernel size from 7 to 5 and eliminated the initial max-pool step. These changes were made to accommodate the reduced size of the imagenet training images. We train for 125 epochs using stochastic gradient descent with momentum of 0.9, weight decay equal to 0.0001, a batch size of 256, and an initial learning rate of 0.1 that decreases by a factor of 10 every 25 epochs. During pretraining, the training dataset was augmented with random crops, random horizontal flips, and random color jitter. The top-1 accuracy after pretraining was 63.9$\%$. For all subsequent training and evaluation steps, the ResNet-18 weights were frozen.The dimensionality of the feature extractor output is $d_f=512$. The hyper-parameters used were derived from the PyTorch \citep{paszke2017automatic} ResNet training tutorial. The only tuning that was performed was on the number of epochs used for training and the interval at which the learning rate was decreased. For the number of epochs, we tried both 90 and 125 epochs and selected 125, which resulted in slightly higher accuracy. We also found that dropping the learning rate at an interval of 25 versus 30 epochs resulted in slightly higher accuracy. 

\subsubsection{Episodic Training of \texorpdfstring{$\vphi$}{TEXT}}
\label{app:hypernet_meta_training}
Next we train the functions that generate the parameters $\vpsi^\tau_f$, $\vpsi^\tau_w$ for the feature extractor adapters and the linear classifier, respectively. We train two variants of \cnaps{} (on ImageNet ILSVRC-2012 only and all datasets - see \cref{table:datasets}). We generate training and validation episodes using the reader from \citep{Triantafillou2019code}. We train in an end-to-end fashion for 110,000 episodes with the Adam \citep{kingma2014adam} optimizer, using a batch size of 16 episodes, and a fixed learning rate of 0.0005. We validate using 200 episodes per validation dataset. Note that when training on ILSVRC only, we validate on ILSVRC only, however, when training on all datasets, we validate on all datasets that have validation data (see \cref{table:datasets}) and consider a model to be better if more than half of the datasets have a higher classification accuracy than the current best model. No data augmentation was employed during the training of $\vphi$. Note that while training $\vphi$ the feature extractor $f_{\vtheta}(\cdot)$ is in `eval' mode (i.e. it will use the fixed batch normalization statistics learned during pretraining the feature extractor weights $\vtheta$ with a moving average). No batch normalization is used in any of the functions generating the $\vpsi^\tau$ parameters, with the exception of the set encoder $g$ (that generates the global task representation $ \vz_\text{G}^\tau $). Note that the target points are never passed through the set encoder $g$. Again, very little hyper-parameter tuning was performed. No grid search or other hyper-parameter search was used. For learning rate we tried both 0.0001 and 0.0005, and selected the latter. We experimented with the number of training episodes in the range of 80,000 to 140,000, with 110,000 episodes generally yielding the best results. We also tried lowering the batch size to 8, but that led to decreased accuracy.

\subsubsection{Evaluation}
\label{app:evaluation}
 We generate test episodes using the reader from \citep{Triantafillou2019code}. We test all models with 600 episodes each on all test datasets. The classification accuracy is averaged over the episodes and a 95\% confidence interval is computed. We compare the best validation and fully trained models in terms of accuracy and use the best of the two. Note that during evaluation, the feature extractor $f_{\vtheta}(\cdot)$ is also in `eval' mode.

\begin{table}[h]
    \caption{Datasets used to train, validate, and test models.}
    \label{table:datasets}
    \tiny
	\centering
	\begin{tabular}{lll|lll}
		\toprule
		\multicolumn{3}{c}{ImageNet ILSVRC-2012} & \multicolumn{3}{c}{All Datasets}\\
		\midrule
		\textbf{Train} & \textbf{Validation} & \textbf{Test} & \textbf{Train} & \textbf{Validation} & \textbf{Test} \\
        \midrule
        ILSVRC \citep{russakovsky2015imagenet} & ILSVRC \citep{russakovsky2015imagenet} & ILSVRC \citep{russakovsky2015imagenet}    & ILSVRC \citep{russakovsky2015imagenet}    & ILSVRC \citep{russakovsky2015imagenet}    & ILSVRC \citep{russakovsky2015imagenet}\\
               &        & Omniglot \citep{lake2011one}  & Omniglot  \citep{lake2011one}   & Omniglot  \citep{lake2011one}  & Omniglot  \citep{lake2011one} \\
               &        & Aircraft \citep{maji2013fine}  & Aircraft \citep{maji2013fine}  & Aircraft \citep{maji2013fine}  & Aircraft \citep{maji2013fine} \\
               &        & Birds \citep{wah2011caltech}     & Birds  \citep{wah2011caltech}    & Birds \citep{wah2011caltech}     & Birds \citep{wah2011caltech} \\
               &        & Textures \citep{cimpoi2014describing}  & Textures \citep{cimpoi2014describing}  & Textures \citep{cimpoi2014describing}  & Textures \citep{cimpoi2014describing} \\
               &        & Quick Draw \citep{ha2017neural} & Quick Draw \citep{ha2017neural} & Quick Draw \citep{ha2017neural} & Quick Draw \citep{ha2017neural} \\
               &        & Fungi \citep{Schroeder2018FGVCx}     & Fungi \citep{Schroeder2018FGVCx}     & Fungi \citep{Schroeder2018FGVCx}     & Fungi \citep{Schroeder2018FGVCx} \\
               &        & VGG Flower \citep{nilsback2008automated} & VGG Flower \citep{nilsback2008automated} & VGG Flower \citep{nilsback2008automated} & VGG Flower \citep{nilsback2008automated} \\
               &        & MSCOCO \citep{lin2014microsoft}    &            & MSCOCO \citep{lin2014microsoft}    & MSCOCO \citep{lin2014microsoft} \\
               &        & Traffic Signs \citep{houben2013detection} &         &            & Traffic Signs \citep{houben2013detection} \\
               &        & MNIST   \citep{lecun2010mnist}   &            &               & MNIST \citep{lecun2010mnist} \\
               &        & CIFAR10  \cite{krizhevsky2009learning}  &            &               & CIFAR10 \cite{krizhevsky2009learning} \\
               &        & CIFAR100 \cite{krizhevsky2009learning}  &            &               & CIFAR100 \cite{krizhevsky2009learning} \\        \bottomrule
	\end{tabular}
    \vspace{2mm}
\end{table}


\section{Additional Few-Shot Classification Results}
\label{app:additional_results}

\subsection{Few-Shot Classification Results When Training on ILSVRC-2012 only}
\label{app:meta_dataset_imagenet}

\cref{table:meta_dataset_results_imagenet} shows few-shot classification results on \textsc{Meta-Dataset} when trained on ILSVRC-2012 only. We emphasize that this scenario does not capture the key focus of our work, and that these results are provided mainly for completeness and compatibility with the work of \citet{triantafillou2019meta}. In particular, our method relies on training the parameters $\vphi$ to adapt the conditional predictive distribution to new datasets. In this setting, the model is never presented with data that has not been used to pre-train $\vtheta$, and therefore cannot learn to appropriately adapt the network to new datasets. Despite this, \cnaps{} demonstrate competitive results with the methods evaluated by \citet{triantafillou2019meta} even in this scenario.

\begin{table}[t]
\caption{Few-shot classification results on \textsc{Meta-Dataset} \citep{triantafillou2019meta} using models trained on ILSVRC-2012 only. All figures are percentages and the $\pm$ sign indicates the 95\% confidence interval. Bold text indicates the highest scores that overlap in their confidence intervals. Results from competitive methods from \citep{triantafillou2019meta}}
\label{table:meta_dataset_results_imagenet}
\small
\centering
\begin{tabular}{@{}lcccccc@{}}
\toprule
Dataset                                   & Finetune              & MatchingNet  & ProtoNet              & fo-MAML      & Proto-MAML            & \cnaps                \\
\midrule
ILSVRC \citep{russakovsky2015imagenet}    & 45.8$\pm$1.1          & 45.0$\pm$1.1 & \textbf{50.5$\pm$1.1} & 36.1$\pm$1.0 & \textbf{51.0$\pm$1.1} & \textbf{50.6$\pm$1.1} \\
\hdashline
Omniglot \citep{lake2011one}              & \textbf{60.9$\pm$1.6}          & 52.3$\pm$1.3 & 60.0$\pm$1.4          & 38.7$\pm$1.4 & \textbf{63.0$\pm$1.4} & 45.2$\pm$1.4          \\
Aircraft \citep{maji2013fine}             & \textbf{68.7$\pm$1.3} & 49.0$\pm$0.9 & 53.1$\pm$1.0          & 34.5$\pm$0.9 & 55.3$\pm$1.0          & 36.0$\pm$0.8          \\
Birds \citep{wah2011caltech}              & 57.3$\pm$1.3          & 62.2$\pm$1.0 & \textbf{68.8$\pm$1.0} & 49.1$\pm$1.2 & \textbf{66.9$\pm$1.0} & 60.7$\pm$0.9          \\
Textures \citep{cimpoi2014describing}     & \textbf{69.1$\pm$0.9} & 64.2$\pm$0.9 & 66.6$\pm$0.8          & 56.5$\pm$0.8 & \textbf{67.8$\pm$0.8} & \textbf{67.5$\pm$0.7} \\
Quick Draw \citep{ha2017neural}           & 42.6$\pm$1.2          & 42.9$\pm$1.1 & 49.0$\pm$1.1          & 27.2$\pm$1.2 & \textbf{53.7$\pm$1.1} & 42.3$\pm$1.0          \\
Fungi \citep{Schroeder2018FGVCx}          & \textbf{38.2$\pm$1.0} & 34.0$\pm$1.0 & \textbf{39.7$\pm$1.1} & 23.5$\pm$1.0 & \textbf{38.0$\pm$1.1} & 30.1$\pm$0.9          \\
VGG Flower \citep{nilsback2008automated}  & \textbf{85.5$\pm$0.7} & 80.1$\pm$0.7 & \textbf{85.3$\pm$0.8} & 66.4$\pm$1.0 & \textbf{86.9$\pm$0.8} & 70.7$\pm$0.7          \\
Traffic Signs \citep{houben2013detection} & \textbf{66.8$\pm$1.3} & 47.8$\pm$1.1 & 47.1$\pm$1.1          & 33.2$\pm$1.3 & 51.2$\pm$1.1          & 53.3$\pm$0.9          \\
MSCOCO \citep{lin2014microsoft}           & 34.9$\pm$1.0          & 35.0$\pm$1.0 & 41.0$\pm$1.1          & 27.5$\pm$1.1 & \textbf{43.4$\pm$1.1} & \textbf{45.2$\pm$1.1} \\
MNIST \citep{lecun2010mnist}              &                       &              &                       &              &                       & \textbf{70.4$\pm$0.8} \\
CIFAR10 \citep{krizhevsky2009learning}    &                       &              &                       &              &                       & \textbf{65.2$\pm$0.8} \\
CIFAR100 \citep{krizhevsky2009learning}   &                       &              &                       &              &                       & \textbf{53.6$\pm$1.0} \\
\bottomrule
\end{tabular}
\end{table}

\subsection{Feature Extractor Parameter Learning}
\label{app:film_visualization}

\begin{figure}[ht]

\centering
\includegraphics[width=1.0\textwidth]{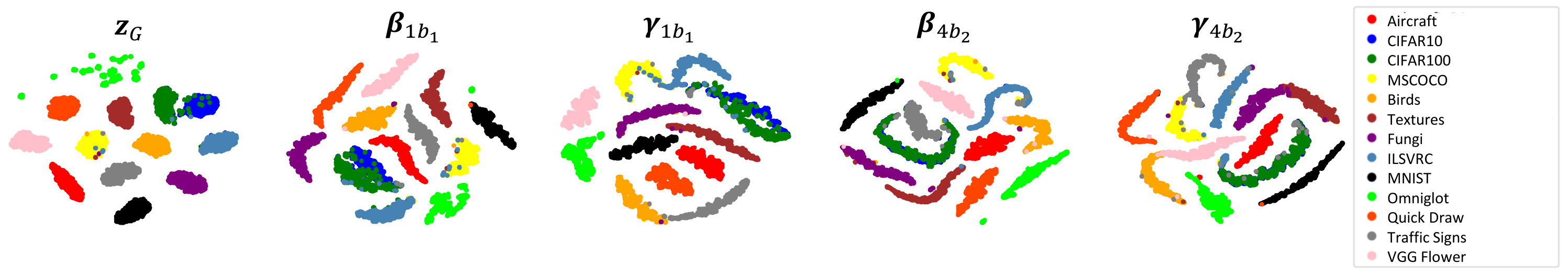} \hfill
\caption{t-SNE plots of the output of the set encoder $\vz_\text{G}$ and the FiLM layer parameters at the start (${\vbeta_{1b1}, \vgamma_{1b1}}$) and end (${\vbeta_{4b2}, \vgamma_{4b2}}$) of the feature extraction process at test time.}
\label{fig:adaptation_tsne}

\end{figure}

\cref{fig:adaptation_tsne} shows t-SNE \citep{maaten2008visualizing} plots that visualize the output of the set encoder $\vz_\text{G}$ and the FiLM layer parameters following the first and last convolutional layers of the feature extractor at test time. Even with unseen test data, the set encoder has learned to clearly separate examples arising from diverse datasets. The FiLM generators learn to generate feature extractor adaptation parameters unique to each dataset. The only significant overlap in the FiLM parameter plots is between CIFAR10 and CIFAR100 datasets which are closely related.

\subsection{Joint Training of \texorpdfstring{$\vtheta$}{TEXT} and \textbf{\texorpdfstring{$\vphi$}{TEXT}} }
\label{app:training_experiments}

Our experiments in jointly training $\vtheta$ and $\vphi$ show that the two-stage training procedure proposed in \cref{sec:training} is crucially important. In particular, we found that joint training diverged in almost all cases we attempted. We were only able to train jointly in two circumstances:
\begin{inlinelist}
\item Using batch normalization in ``train'' mode for both context \textit{and} target sets. We stress that this implies computing the batch statistics at test time, and using those to normalize the batches. This is in contrast to the methodology we propose in the main text: only using batch normalization in ``eval'' mode, which enforces that no information is transferred across tasks or datasets.
\item ``Warm-start" the training procedure with batch normalization in ``train'' mode, and after a number of epochs (we use 50 for the results shown below), switch to proper usage of batch normalization.
\end{inlinelist}
All other training procedures we attempted diverged.
\begin{table}[]
\caption{Few-shot classification results on \textsc{Meta-Dataset} \citep{triantafillou2019meta} comparing joint training for $\vtheta$ and $\vphi$ (columns 2 and 3) to two-stage training (column 4). All figures are percentages and the $\pm$ sign indicates the 95\% confidence interval. Bold text indicates the highest scores that overlap in their confidence intervals.}
\label{table:training_comparison}
\begin{tabular}{@{}lccc@{}}
\toprule
Dataset & \begin{tabular}[c]{@{}c@{}}Joint Training\\ (warmstart BN)\end{tabular} & \begin{tabular}[c]{@{}c@{}}Joint Training\\ (BN train mode)\end{tabular} & \begin{tabular}[c]{@{}c@{}}Two-Stage Training\\ (BN test mode)\end{tabular} \\ \midrule
ILSVRC \citep{russakovsky2015imagenet} & 17.3$\pm$0.7 & 41.6$\pm$1.0 & 49.5$\pm$1.0 \\
Omniglot \citep{lake2011one} & 74.9$\pm$1.0 & 80.8$\pm$0.9 & 89.7$\pm$0.5 \\
Aircraft \citep{maji2013fine} & 51.4$\pm$0.8 & 70.5$\pm$0.7 & 87.2$\pm$0.5 \\
Birds \citep{wah2011caltech} & 44.1$\pm$1.0 & 48.3$\pm$1.0 & 76.7$\pm$0.9 \\
Textures \citep{cimpoi2014describing} & 49.1$\pm$0.7 & 73.5$\pm$0.6 & 83.0$\pm$0.6 \\
Quick Draw \citep{ha2017neural} & 46.6$\pm$1.0 & 71.5$\pm$0.8 & 72.3$\pm$0.8 \\
Fungi \citep{Schroeder2018FGVCx} & 20.4$\pm$0.9 & 43.1$\pm$1.1 & 50.5$\pm$1.1 \\
VGG Flower \citep{nilsback2008automated} & 66.6$\pm$0.8 & 71.0$\pm$0.7 & 92.5$\pm$0.4 \\
\hdashline
Traffic Signs \citep{houben2013detection} & 21.2$\pm$0.8 & 40.4$\pm$1.1 & 48.4$\pm$1.1 \\
MSCOCO \citep{lin2014microsoft} & 18.8$\pm$0.7 & 37.1$\pm$1.0 & 39.7$\pm$0.9 \\ \bottomrule
\end{tabular}
\end{table}

\cref{table:training_comparison} details the results of our study on training procedures. The results demonstrate that the two-stage greatly improves performance of the model, even compared to using batch normalization in ``train mode'', which gives the model an unfair advantage over our standard model.

\subsection{Comparison Between \cnaps{} and Parallel Residual Adapters \citep{rebuffi2018efficient}}
\label{app:parallel_residual_adapter_comparison}
 \cnaps{} adds FiLM layers \citep{perez2018film} in series with each convolutional layer to adapt the feature extractor to a particular task while parallel residual adapters from \citet{rebuffi2018efficient} adds $1 \times 1$ convolutions in parallel with each convolution layer to do the same. However, if the number of feature channels is $C$, then the number of parameters required for each convolutional layer in the feature extractor is $2C$ for \cnaps{} and $C^2$ for parallel residual adapters. Hence, parallel residual adapters have $C/2$ times the capacity compared to FiLM layers. Despite this advantage, CNAPs achieves superior results as can be seen in \cref{table:parallel_residual_adapter_comparison}.

\begin{table}[t]
    \caption{Few-shot classification results on \textsc{Meta-Dataset} \citep{triantafillou2019meta} using models trained on all training datasets for Parallel Residual Adapters \citep{rebuffi2018efficient} and \cnaps{}. All figures are percentages and the $\pm$ sign indicates the 95\% confidence interval over tasks. Bold text indicates the scores within the confidence interval of the highest score. Tasks from datasets below the dashed line were not used for training.}
    \label{table:parallel_residual_adapter_comparison}
	\centering
	\begin{tabular}{lcc}
	\toprule
	Dataset                                   & Parallel Residual Adapter & \cnaps{} \\
    \midrule
	ILSVRC \citep{russakovsky2015imagenet}    & \textbf{51.2 $\pm$ 1.0}   & \textbf{52.3 $\pm$ 1.0} \\
	Omniglot \citep{lake2011one}              & \textbf{87.3 $\pm$ 0.7}   & \textbf{88.4 $\pm$ 0.7} \\
    Aircraft \citep{maji2013fine}             & 78.3 $\pm$ 0.7            & \textbf{80.5 $\pm$ 0.6} \\
    Birds \citep{wah2011caltech}              & 67.8 $\pm$ 0.9            & \textbf{72.2 $\pm$ 0.9} \\
    Textures \citep{cimpoi2014describing}     & 55.5 $\pm$ 0.7            & \textbf{58.3 $\pm$ 0.7} \\
    Quick Draw \citep{ha2017neural}           & 70.9 $\pm$ 0.7            & \textbf{72.5 $\pm$ 0.8} \\
    Fungi \citep{Schroeder2018FGVCx}          & 44.6 $\pm$ 1.1            & \textbf{47.4 $\pm$ 1.0} \\
    VGG Flower \citep{nilsback2008automated}  & 81.7 $\pm$ 0.7            & \textbf{86.0 $\pm$ 0.5} \\
	\hdashline
    Traffic Signs \citep{houben2013detection} & 57.2 $\pm$ 0.9            & \textbf{60.2 $\pm$ 0.9} \\ 
    MSCOCO \citep{lin2014microsoft}           & \textbf{43.7 $\pm$ 1.0}   & \textbf{42.6 $\pm$ 1.1} \\
    MNIST \citep{lecun2010mnist}              & 91.1 $\pm$ 0.4            & \textbf{92.7 $\pm$ 0.4} \\
    CIFAR10 \citep{krizhevsky2009learning}    & \textbf{64.5 $\pm$ 0.8}   &         61.5 $\pm$ 0.7 \\ 
    CIFAR100 \citep{krizhevsky2009learning}   & \textbf{50.4 $\pm$ 0.9}   & \textbf{50.1 $\pm$ 1.0} \\
    \bottomrule
\end{tabular}
\end{table}

\section{Network Architecture Details}
\label{app:network_architecture_details}

\subsection{ResNet18 Architecture details}
\label{app:resnet_architecture}

Throughout our experiments in \cref{sec:experiments}, we use a ResNet18 \citep{he2016deep} as our feature extractor, the parameters of which we denote $\vtheta$. \cref{table:basic_resnet_block} and \cref{table:basic_resnet_scaling_block} detail the architectures of the basic block (left) and basic scaling block (right) that are the fundamental components of the ResNet that we employ. \cref{table:pre_trained_resnet_architecture} details how these blocks are composed to generate the overall feature extractor network. We use the implementation that is provided by the PyTorch \citep{paszke2017automatic}\footnote{https://pytorch.org/docs/stable/torchvision/models.html}, though we adapt the code to enable the use of FiLM layers.

\begin{table}[!htb]
    \RawFloats
    \adjustbox{valign=t}{\begin{minipage}{.5\linewidth}
      \caption{ResNet-18 basic block $b$.}
      \label{table:basic_resnet_block}
      \centering
        \begin{tabular}{l}
		\toprule
		\textbf{Layers} \\
        \midrule
        Input \\
        Conv2d ($3 \times 3$, stride 1, pad 1) \\ 
        BatchNorm \\
        FiLM ($\vgamma_{b,1}, \vbeta_{b,1}$) \\
        ReLU \\
        Conv2d ($3 \times 3$, stride 1, pad 1) \\ 
        BatchNorm \\
        FiLM ($\vgamma_{b,2}, \vbeta_{b,2}$) \\
        Sum with Input \\
        ReLU \\
        \bottomrule
	\end{tabular}
    \end{minipage}}%
    \adjustbox{valign=t}{\begin{minipage}{.5\linewidth}
      \centering
        \caption{ResNet-18 basic scaling block $b$.}
        \label{table:basic_resnet_scaling_block}
        \begin{tabular}{l}
		\toprule
		\textbf{Layers} \\
        \midrule
        Input \\
        Conv2d ($3 \times 3$, stride 2, pad 1) \\ 
        BatchNorm \\
        FiLM ($\vgamma_{b,1}, \vbeta_{b,1}$) \\
        ReLU \\
        Conv2d ($3 \times 3$, stride 1, pad 1) \\ 
        BatchNorm \\
        FiLM ($\vgamma_{b,2}, \vbeta_{b,2}$) \\
        Downsample Input by factor of 2 \\
        Sum with Downsampled Input \\
        ReLU \\
        \bottomrule
	\end{tabular}
    \end{minipage}}
\end{table}

\begin{table}[h]
    \caption{ResNet-18 feature extractor network.}
    \label{table:pre_trained_resnet_architecture}
	\centering
	\begin{tabular}{lcl}
		\multicolumn{3}{l}{\textbf{ResNet-18 Feature Extractor ($\theta$) with FiLM Layers:} $\vx \rightarrow f_\vtheta(\vx; \vpsi_{f}^{\tau})$, $\vx^* \rightarrow f_\vtheta(\vx^*; \vpsi_{f}^{\tau})$ } \\
		\toprule
		\textbf{Stage}  & \textbf{Output size}      & \textbf{Layers} \\
        \midrule
        Input           & $84 \times 84 \times 3$   & Input image \\
        Pre-processing  & $41 \times 41 \times 64$  & Conv2d ($5 \times 5$, stride 2, pad 1, BatchNorm, ReLU) \\
        Layer $1$         & $41 \times 41 \times 64$  & Basic Block $\times$ 2 \\
        Layer $2$         & $21 \times 21 \times 128$ & Basic Block, Basic Scaling Block \\
        Layer $3$         & $11 \times 11 \times 256$ & Basic Block, Basic Scaling Block \\
        Layer $4$         & $6 \times 6 \times 512$   & Basic Block, Basic Scaling Block \\
        Post-Processing & 512                       & AvgPool, Flatten \\
        \bottomrule
	\end{tabular}
    \vspace{2mm}
\end{table}

\subsection{Adaptation Network Architecture Details}
\label{app:adaptation_network_architectures}

In this section, we provide the details of the architectures used for our adaptation networks. \cref{table:global_set_encoder} details the architecture of the set encoder $g:D^\tau \mapsto \vz_{\text{G}} $ that maps context sets to global representations. 
\begin{table}[h]
    \caption{Set encoder $g$.}
    \label{table:global_set_encoder}
	\centering
	\begin{tabular}{cl}
		\multicolumn{2}{l}{\textbf{Set Encoder ($g$):} $\vx \rightarrow {\vz}_G^{\tau}$} \\
		\toprule
		\textbf{Output size} & \textbf{Layers} \\
        \midrule
		$84 \times 84 \times 3$ & Input image \\
		$42 \times 42 \times 64$ & Conv2d ($3 \times 3$, stride 1, pad 1, ReLU), MaxPool ($2 \times 2 $, stride 2) \\
		$21 \times 21 \times 64$ & Conv2d ($3 \times 3$, stride 1, pad 1, ReLU), MaxPool ($2 \times 2 $, stride 2) \\
		$10 \times 10 \times 64$ & Conv2d ($3 \times 3$, stride 1, pad 1, ReLU), MaxPool ($2 \times 2 $, stride 2) \\
		$5 \times 5 \times 64$ & Conv2d ($3 \times 3$, stride 1, pad 1, ReLU), MaxPool ($2 \times 2 $, stride 2) \\
		$2 \times 2 \times 64$ & Conv2d ($3 \times 3$, stride 1, pad 1, ReLU),  MaxPool ($2 \times 2 $, stride 2) \\
        64 & AdaptiveAvgPool2d \\
        \bottomrule
	\end{tabular}
\end{table}

\cref{table:ar_set_encoders} details the architecture used in the auto-regressive parameterization of $\vz_{\text{AR}}$. In our experiments, there is one such network for every block in the ResNet18 (detailed in \cref{table:pre_trained_resnet_architecture}). These networks accept as input the set of activations from the previous block, and map them (through the permutation invariant structure) to a vector representation of the output of the layer. The representation $\vz_i = (\vz_{\text{G}}, \vz_{\text{AR}})$ is then generated by concatenating the global and auto-regressive representations, and fed into the adaptation network that provides the FiLM layer parameters for the next layer. This network is detailed in \cref{table:film_generator}, and illustrated in \cref{fig:film_generators}. Note that, as depicted in \cref{fig:film_generators}, each layer has four networks with architectures as detailed in \cref{table:film_generator}, one for each $\vgamma$ and $\vbeta$, for each convolutional layer in the block.

\begin{table}[h]
    \caption{Network of set encoder $\vphi_f$.}
    \label{table:ar_set_encoders}
    \centering
	\begin{tabular}{cl}
	  \multicolumn{2}{l}{\textbf{Set Encoder ($\vphi_f$):} 
	  $\{f_{\vtheta}^{l_i}(x; \vpsi_f^{\tau})\}$ $\rightarrow \vz^i_{\text{AR}}$ } \\
      \toprule
      \textbf{Output size} & \textbf{Layers} \\
      \midrule
       $l_i$ channels $\times$ $l_i$ channel size & Input $\{f_{\vtheta}^{l_i}(x; \vpsi_f^{\tau})\}$ \\
       $l_i$ channels $\times$ $l_i$ channel size & AvgPool, Flatten \\
       $l_i$ channels & fully connected, ReLU \\
       $l_i$ channels & 2 $\times$ fully connected with residual skip connection, ReLU \\
       $l_i$ channels & fully connected with residual skip connection \\
       $l_i$ channels & mean pooling over instances \\
       $l_i$ channels & Input from mean pooling \\
       $l_i$ channels & fully connected, ReLU \\      \bottomrule
	\end{tabular}
\end{table}

\begin{table}[h]
    \caption{Network $\phi_f$.}
    \label{table:film_generator}
	\centering
	\begin{tabular}{cl}
	  \multicolumn{2}{l}{\textbf{Network ($\phi_f$):} $(\vz_{\text{G}}, \vz_{\text{AR}})\rightarrow (\vgamma, \vbeta)$}\\
 		\toprule
        \textbf{Output size} & \textbf{Layers} \\
        \midrule
        $64 + l_i$ channels &  Input from Concatenate \\
        $l_i$ channels & fully connected, ReLU \\
        $l_i$ channels & 2 $\times$ fully connected with residual skip connection, ReLU \\
        $l_i$ channels & fully connected with residual skip connection \\
		\bottomrule
	\end{tabular}
\end{table}

\subsection{Linear Classifier Adaptation Network}
\label{app:linear_classifier}

Finally, in this section we give the details for the linear classifer $\vpsi^\tau_w$, and the adaptation network that provides these task-specific parameters $\vpsi_w(\cdot)$. The adaptation network accepts a class-specific representation that is generated by applying a mean-pooling operation to the adapted feature activations of each instance associated with the class in the context set: $\vz^\tau_c = \frac{1}{N^\tau_c}\sum\limits_{\vx \in D^\tau_c} f_\vtheta(\vx; \vpsi^\tau_f)$, where $N^\tau_c$ denotes the number of context instances associated with class $c$ in task $\tau$. $\vpsi_w$ is comprised of two separate networks (one for the weights  $\vpsi_{w}$ and one for the biases $\vpsi_{b}$) detailed in \cref{table:psi_w_adaptation_networks} and \cref{table:psi_b_adaptation_networks}. The resulting weights and biases (for each class in task $\tau$) can then be used as a linear classification layer, as detailed in \cref{table:linear_classifier}.

\begin{table}[!htb]
    \RawFloats
    \adjustbox{valign=t}{\begin{minipage}{.5\linewidth}
      \caption{Network $\vphi_w$.}
      \label{table:psi_w_adaptation_networks}
      \centering
      \begin{tabular}{cl}
	    \multicolumn{2}{l}{\textbf{Network ($\vphi_w$):}} \\
	    \multicolumn{2}{l}{$\vz_c \rightarrow \vpsi_{w,w}$} \\
        \toprule
        \textbf{Output size} & \textbf{Layers} \\
        \midrule
        $512$ & Input from mean pooling \\
        $512$ & 2 $\times$ fully connected, ELU \\
        $512$ & fully connected \\
        $512$ & Sum with Input \\
        \bottomrule
	  \end{tabular}
    \end{minipage}}%
    \adjustbox{valign=t}{\begin{minipage}{.5\linewidth}
      \centering
      \caption{Network $\vphi_b$.}
      \label{table:psi_b_adaptation_networks}
      \begin{tabular}{cl}
	    \multicolumn{2}{l}{\textbf{Network ($\vphi_b$):}} \\
	    \multicolumn{2}{l}{$\vz_c \rightarrow \vpsi_{w,b}$} \\
        \toprule
        \textbf{Output size} & \textbf{Layers} \\
        \midrule
        $512$ & Input from mean pooling \\
        $512$ & 2 $\times$ fully connected, ELU \\
        $1$ & fully connected \\
        \bottomrule
	  \end{tabular}
    \end{minipage}}
\end{table}

\begin{table}[h]
    \caption{Linear classifier network.}
	\centering
	\begin{tabular}{ll}
		\multicolumn{2}{l}{\textbf{Linear Classifier ($\vpsi_w$):} $f_\vtheta(\vx^\ast; \vpsi_{f}^{\tau}) \rightarrow p(\vy^\ast | \vx^\ast, \vpsi^\tau(D^\tau), \vtheta)$} \\
 		\toprule
        \textbf{Output size} & \textbf{Layers} \\
        \midrule
		$512$ & Input features $f_\vtheta(\vx^\ast; \vpsi_{f}^{\tau})$ \\
		$512 \times C^\tau$ & Input weights $w$ \\
		$512 \times 1$ & Input biases $b$ \\
		$C^\tau$ & fully connected \\
		$C^\tau$ & softmax \\ 
		\bottomrule
	\end{tabular}
	\label{table:linear_classifier}
\end{table}

\clearpage
\newpage

\section{Continual Learning Implementation Details}
\label{app:additional_continual_learning_details}

As noted in \cref{sec:model,sec:experiments}, our model can be applied to continual learning with one small modification: we store a compact representation of our training data that can be updated at each step of the continual learning procedure. Notice that \cref{fig:classifier} indicates that the functional representation of our linear classification layer $\vpsi^\tau_w(\cdot)$ contains a mean pooling layer that combines the per-class output of our feature extractor $\{ f_\vtheta \left( \vx_m^\tau; \vpsi_f \right) | \vx_m^\tau \in D^\tau, \vy^\tau_m = c\}$. The result of this pooling, 
\begin{equation}
\vz_c = \frac{1}{M} \sum f_\vtheta \left( \vx_m^\tau; \vpsi_f \right)
\end{equation}
where $M = |\{ f_\vtheta \left( \vx_m^\tau; \vpsi_f \right) | \vx_m^\tau \in D^\tau, \vy^\tau_m = c\}|$, is supplied as input to the network $\vpsi_w(\cdot)$. This network yields the class conditional parameters of the linear classifier $\vpsi^\tau_w$, resulting in (along with the feature extractor parameters $\vpsi^\tau_f$) the full paramterization of $\vpsi^\tau$. We store $\vz_c$ as the training dataset representation for, class $c$. 

If at any point in our continual learning procedure we observe new training data for class $c$ we can update our representation for class $c$ by computing 
$\vz_c' = \frac{1}{M} \sum f_\vtheta \left( {\vx_m^{\tau}}'; \vpsi_f \right)$ the pooled average resulting from $M$ new training examples ${\vx_m^{\tau}}'$ for class $c$. We then update $\vz_c$ with the weighted average:
$\vz_c \leftarrow \frac{M \vz_c + N \vz_c}{M+N}$. At prediction time, we supply $\vz_c$ to $\vpsi_w(\cdot)$ to produce classification parameters for class $c$. 

Similar to the input to $\vpsi^\tau_w(\cdot)$, the input to $\vpsi^\tau_f(\cdot)$ also contains a mean-pooled representation, this time of the entire training dataset $\vz^\tau_G$. This representation is also stored and updated in the same way.

One issue with our procedure is that it is not completely invariant to the order in which we observe the sequence of training data during our continual learning procedure. The feature extractor adaptation parameters are only conditioned on the most recent training data, meaning that if data from class $c$ is not present in the most recent training data, $z_c$ was generated using "old" feature extractor adaptation parameters (from a previous time step). This creates a potential disconnect between the classification parameters from previous time steps and the feature extractor output. Fortunately, in our experiment we noticed little within dataset variance for the adaptation parameters. Since all of our experiments on continual learning were within a single dataset, this did not seem to be an issue as \cnaps{} were able to achieved good performance. However, for continual learning experiments that contain multiple datasets, we anticipate that this issue will need to be addressed.

\section{Additional Continual Learning Results}
\label{app:additional_continual_learning_results}

In \cref{sec:experiments} we provided results for continual learning experiments with Split MNIST \citep{zenke2017continual} and Split CIFAR100 \citep{chaudhry2018riemannian}. The results showed the average performance as more tasks were observed for the single and multi head settings. Here, we provide more complete results, detailing the performance through ``time" at the task level. 
\begin{figure}[htb]
     \includegraphics[width=\linewidth]{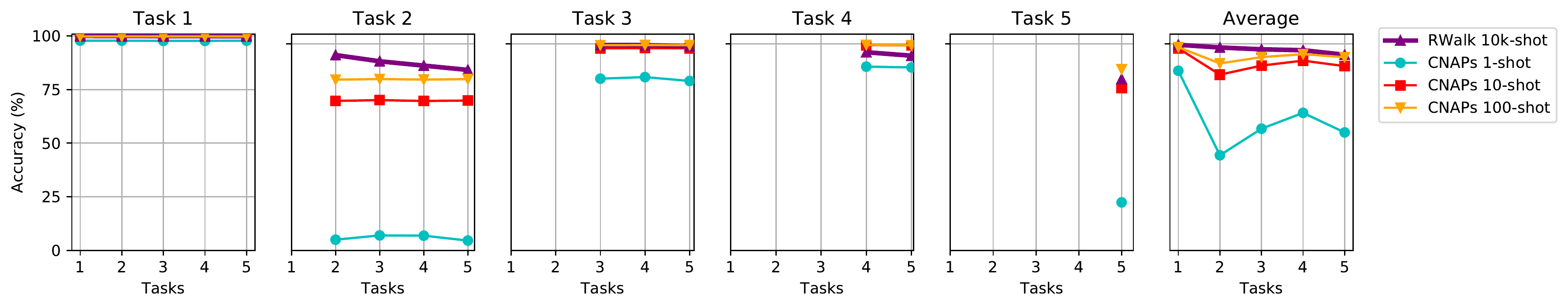} \\
    \includegraphics[width=\linewidth]{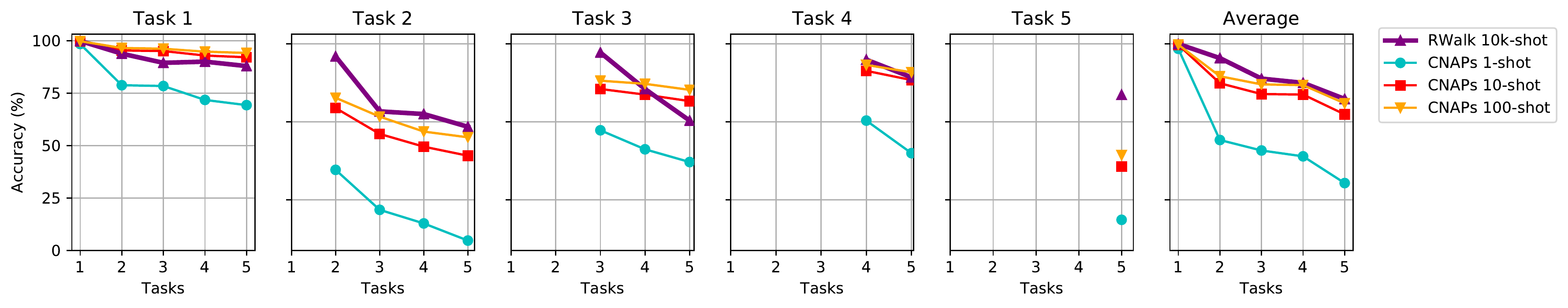} \\
    \caption{Continual learning results on Split MNIST. Top row is multi-head, bottom row is single-head.}
    \label{fig:continual_learning_mnist}
\end{figure}
\cref{fig:continual_learning_mnist} details the performance of \cnaps{} (with varying number of observed examples) and Riemannian Walk (RWalk) \citep{chaudhry2018riemannian} on the five tasks of Split MNIST through time. Note that RWalk makes explicit use of training data from previous time steps when new data is observed, while \cnaps{} do not. 

\cref{fig:continual_learning_mnist} implies that \cnaps{} is competitive with RWalk in this scenario, despite seeing far less data per task, and not using old data to retrain the model at every time-step. Further, we see that \cnaps{} is naturally resistant to forgetting, as it uses internal task representations to maintain important information about tasks seen at previous time-steps.

\begin{figure}[htb]
    \includegraphics[width=\linewidth]{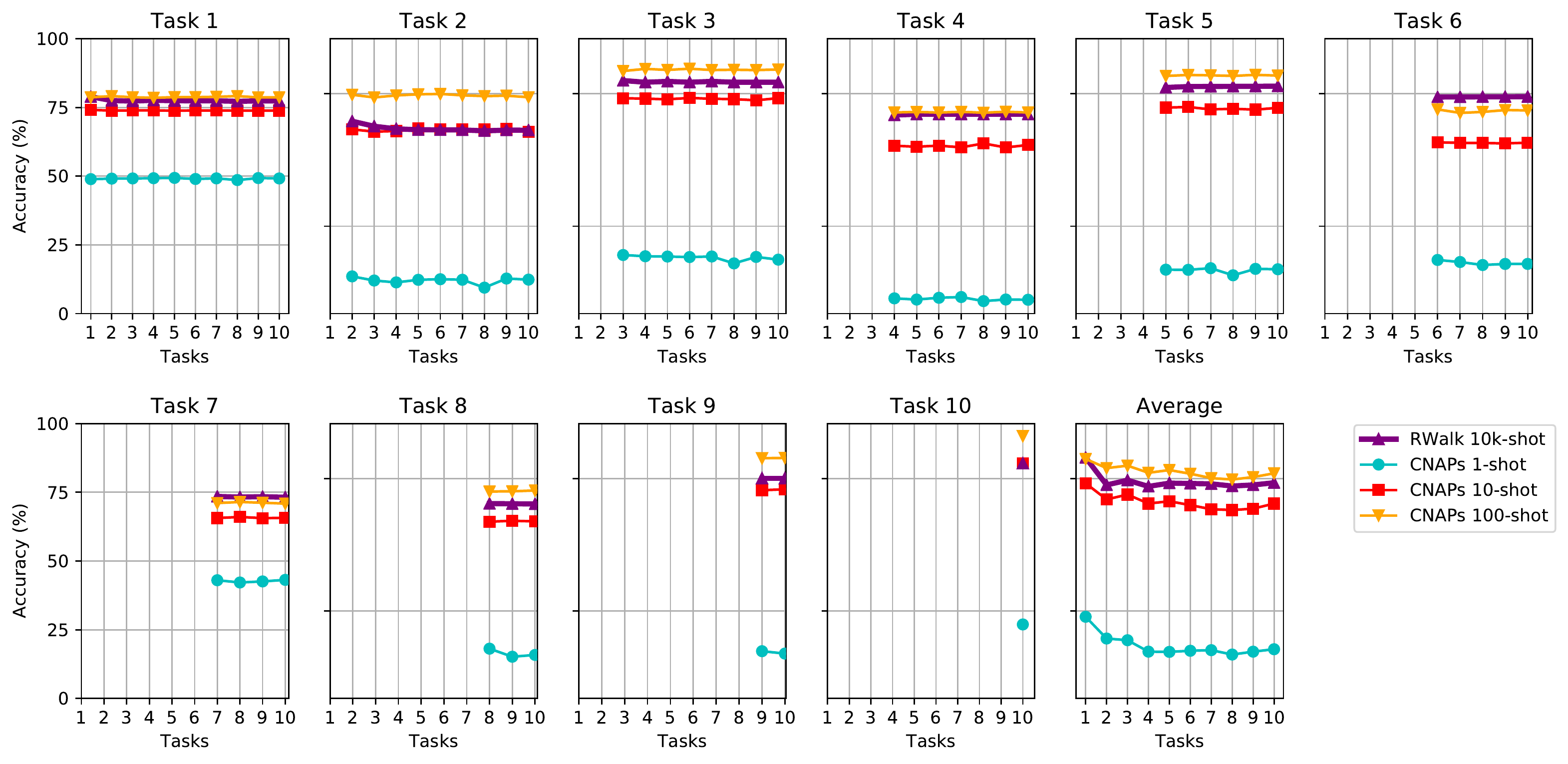} \\
    \includegraphics[width=\linewidth]{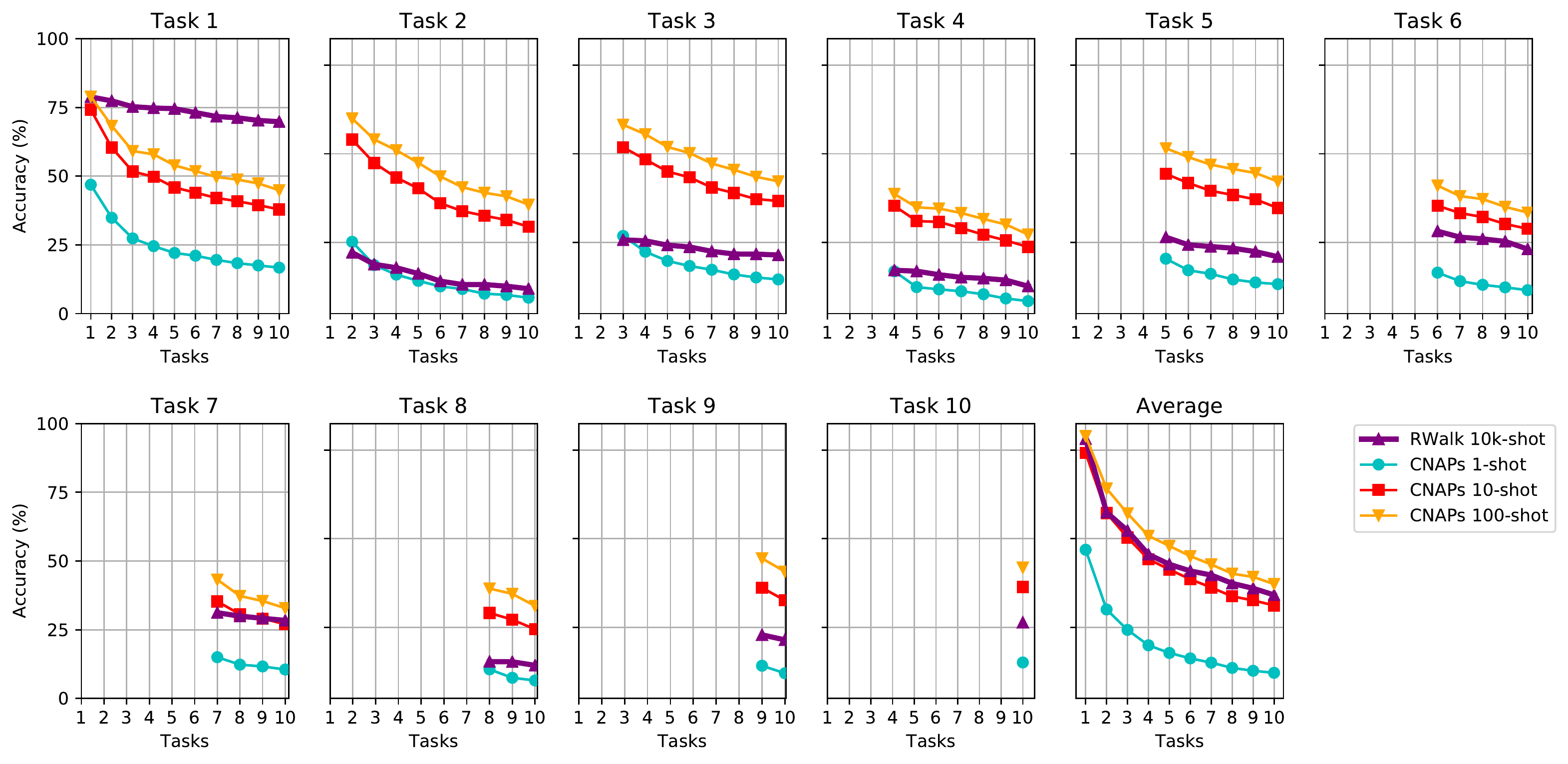} \\
    \caption{Continual learning results on Split CIFAR100. Top two rows are multi-head, bottom two rows are single-head.}
    \label{fig:continual_learning_cifar100}
\end{figure}

\cref{fig:continual_learning_cifar100} demonstrates that \cnaps{} maintains similar results when scaling up to considerably more difficult datasets such as CIFAR100. Here too, \cnaps{} has not been trained on this dataset, yet demonstrates performance comparable to (and even better than) RWalk, a method explicitly trained for this task that makes use of samples from previous tasks at each time step. 

\clearpage
\newpage

\section{Additional Active Learning Results}
\label{app:additional_active_learning_results}

In \cref{sec:experiments} we provided active learning results for \cnaps{} and 
Prototypical Networks on the VGG Flowers dataset and three held out test languages from the Omniglot dataset. Here, we provide the results from all twenty held-out languages in Omniglot.

\begin{figure}[htb]
     \includegraphics[width=\linewidth]{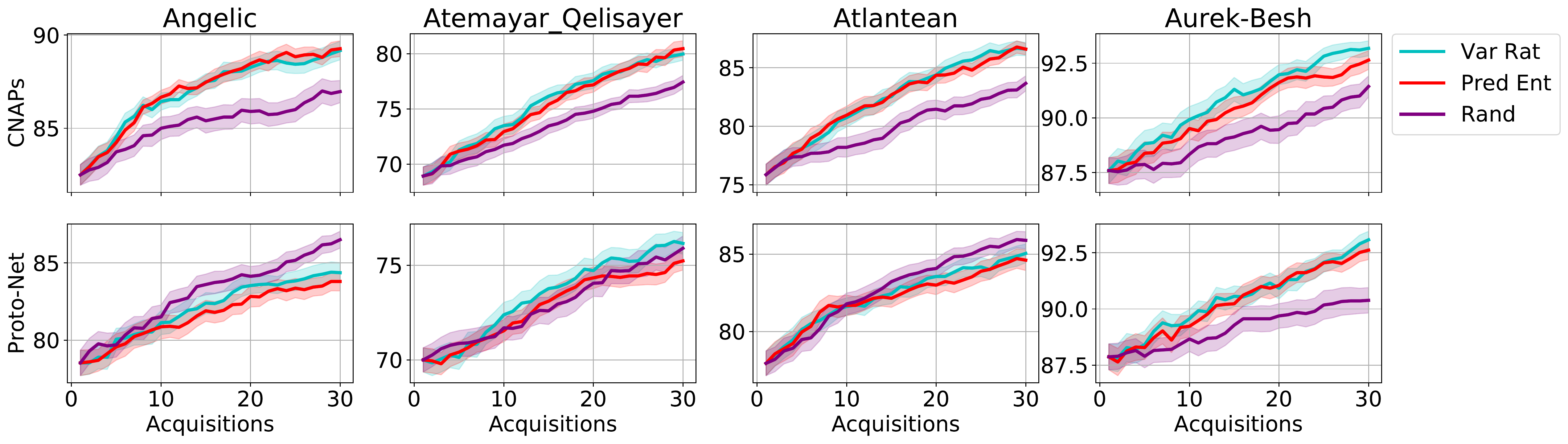} \\
     \includegraphics[width=\linewidth]{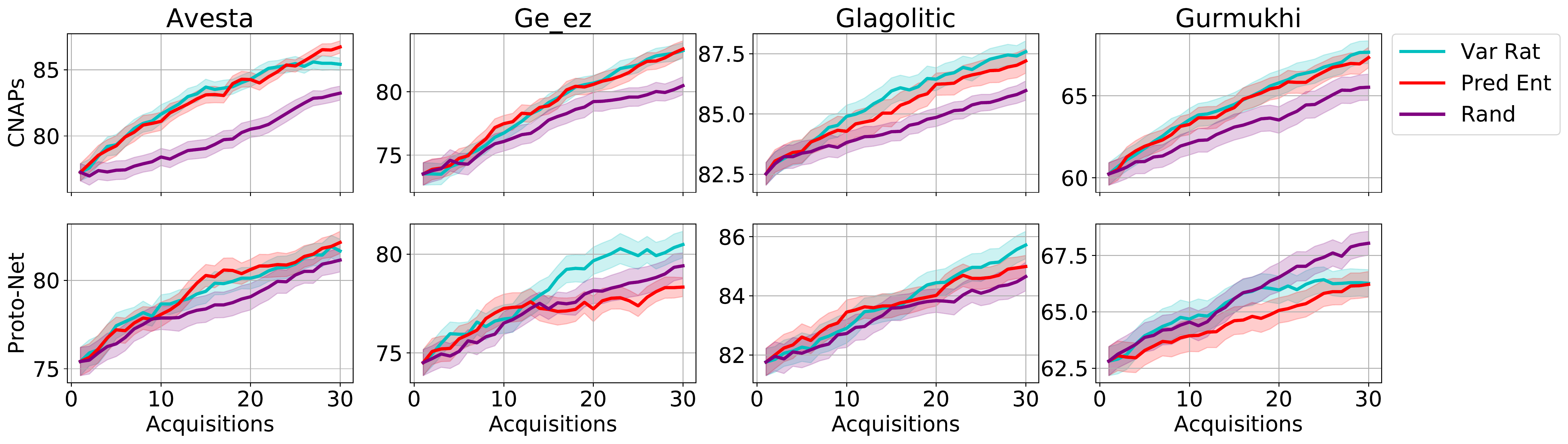} \\
     \includegraphics[width=\linewidth]{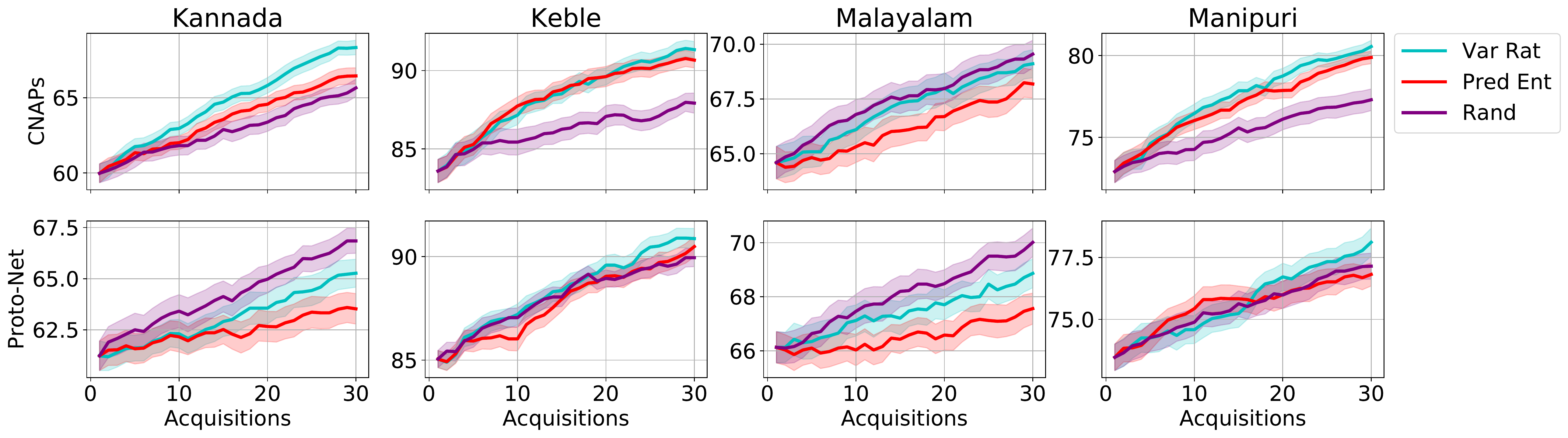} \\
     \includegraphics[width=\linewidth]{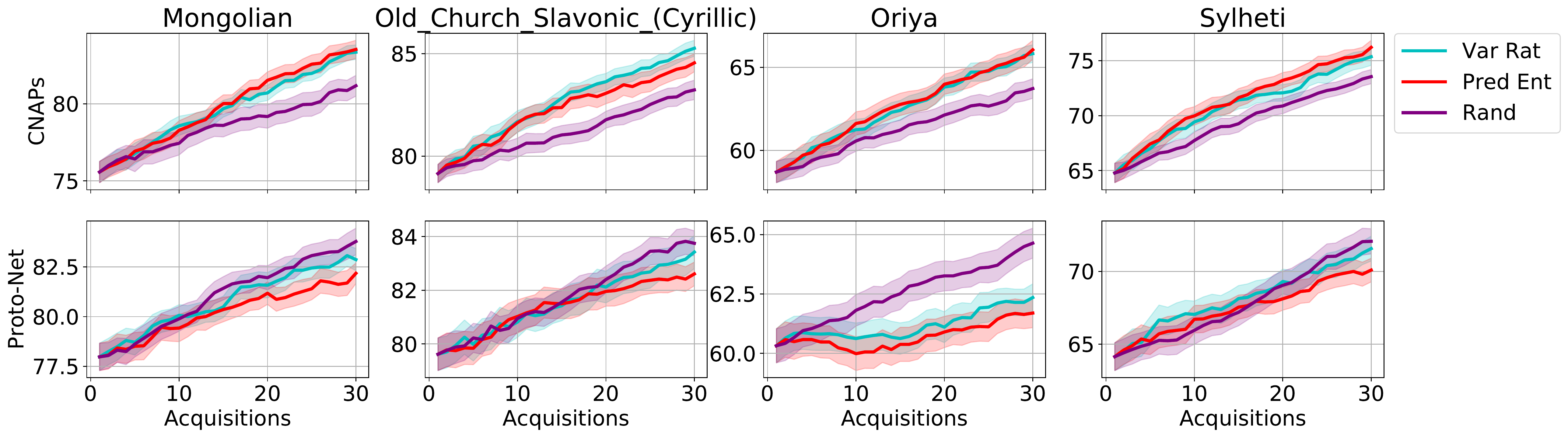} \\
     \includegraphics[width=\linewidth]{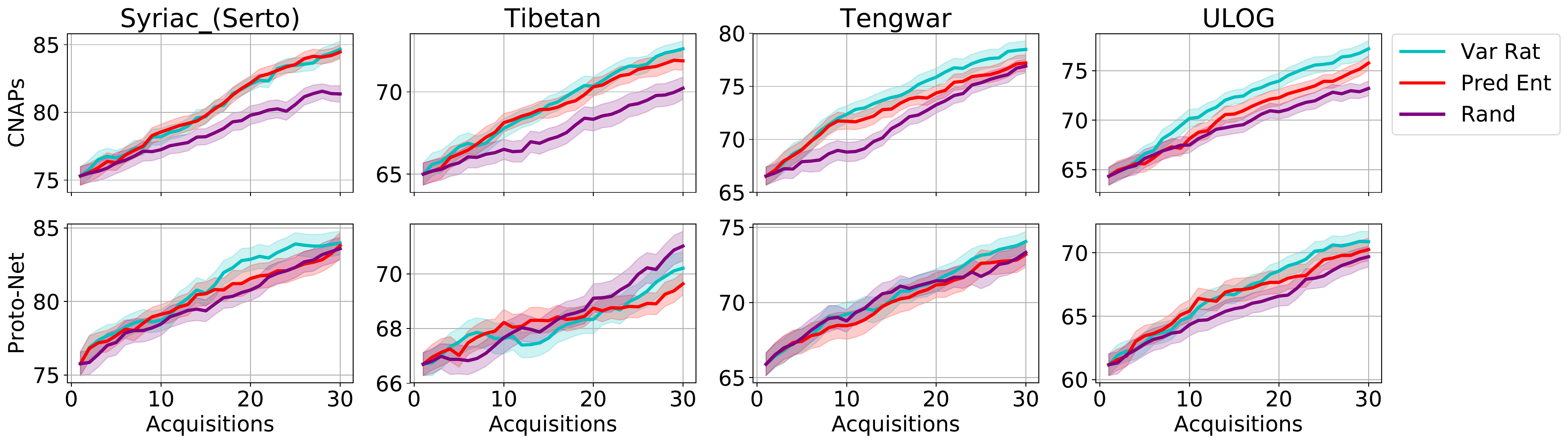} \\
     \caption{Active learning results on all twenty held-out \textsc{Omniglot} languages.}
     \label{fig:active_learning_complete}
\end{figure}

\cref{fig:active_learning_complete} demonstrates that in almost all held-out languages, using the predictive distribution of \cnaps{} not only improves overall performance, but also enables the model to make use of standard acquisition functions \citep{cohn1996active} to improve data efficiency over random acquisition. In contrast, we see that in most cases, random acquisition performs as well or better than acquisition functions that rely on the predictive distribution of Prototypical Networks. This provides empirical evidence that in addition to achieving overall better performance, the predictive distribution of \cnaps{} is more calibrated, and thus better suited to tasks such as active learning that require uncertainty in predictions.


\clearpage
\newpage


\end{document}